\documentclass[11pt]{article}
\usepackage{geometry}
\usepackage{makeidx}
\usepackage{graphicx}
\usepackage{amsmath,amssymb} 
\usepackage{color}
\usepackage[utf8]{inputenc}
\usepackage{graphicx}
\graphicspath{{Figures/}} 
\usepackage{caption}

\usepackage{amsthm}
\usepackage{float}
\usepackage{authblk}
\usepackage[linesnumbered,ruled]{algorithm2e}
\usepackage{hyperref} 

\DeclareMathOperator{\E}{\mathbb{E}}

\newtheorem{prop}{Proposition}
\title{\textbf{Texture and Structure Two-view Classification of Images}}

\author{Samah Khawaled$^1$\thanks{ssamahkh@campus.technion.ac.il}, Michael Zibulevsky$^2$, Yehoshua Y. Zeevi$^1$}
\affil{ $^1$Electrical Engineering, Technion Institute of Technology, Israel\\
$^2$Computer Science, Technion Institute of Technology, Israel}
\date{}
\begin{document}

	\maketitle
	\begin{abstract}
	Textural and structural features can be regraded as "two-view" feature sets. Inspired by the recent progress in multi-view learning, we propose a novel two-view classification method that models each feature set and optimizes the process of merging these views efficiently. Examples of implementation of this approach in classification of real-world data are presented, with special emphasis on medical images. We firstly decompose fully-textured images into two layers of representation, corresponding to natural stochastic textures (NST) and structural layer, respectively. The structural, edge-and-curve-type, information is mostly represented by the local spatial phase, whereas, the pure NST has random phase and is characterized by Gaussianity and self-similarity. Therefore, the NST is modeled by the 2D self-similar process, fractional Brownian motion (fBm). The Hurst parameter, characteristic of fBm, specifies the roughness or irregularity of the texture. This leads us to its estimation and implementation along other features extracted from the structure layer, to build the "two-view" features sets used in our classification scheme. A shallow neural net (NN) is exploited to execute the process of merging these feature sets, in a straightforward and efficient manner. 
	\end{abstract}

	\section{Introduction}
	\label{sec:intro}
	Texture is one of the key image attributes that are crucial for natural image processing, analysis and understanding. Textures convey information regarding the physical properties of various natural substances, including biological tissue. Textures are generally divided into three classes: regular/periodic, irregular and stochastic \cite{Lin2004}. In contrast with regular textures, which depict a semi-periodic structure, stochastic textures appear to look more like spatial noise, and they are characterized by their statistical properties. Of special interest are the Natural Stochastic Textures (NST). Previous studies have shown that natural structured images exhibit non-Gaussian distribution and are characterized by high Kurtosis, as exhibited for example, by the 1D marginal histograms of the wavelets coefficients \cite{Wainwright}. In contrast with structured images that are dominated by edges and contours, NST images obey Gaussian behavior \cite{ZZ}, and they are characterized by the statistical properties of non-stationarity and self-similarity. In fact, the 2D fractional Brownian Motion (fBm) \cite{Mandel}, a self-similar non-stationary Gaussian process, has been shown to be a suitable model for many NST images \cite{NST1,NST2}.\\
	\indent Most natural fully-textured images incorporate both stochastic textures and structural (edges, contours and thin lines detailed) information, even over segments that look like uniform NST, wherein the structured details are characterized by local phase coherence. We therefore separate the images into a layer of pure NST and a layer of structured information. We then assess the extent to which the texture is fractal in the sense of statistical self-similarity, by measuring the resemblance between different scales of its multi-resolution wavelets coefficients. This can be done by examining the invariance along scales \cite{abry2015irregularities}. Under the assumption that the covariance obeys the fBm statistics, we proceed to estimate the Hurst parameter of the fBm, presumed to underlie the process \cite{Mandel}.\\
	\indent Multi-view learning, also known as data fusing, is capable of boosting the generalization performance by learning each view separately \cite{zhao2017multi}. Multi-view supervised algorithms, based on Support Vector Machine (SVM) were recently proposed \cite{sun2011multi}. Their applications in detection, recognition and classification are discussed in \cite{nasiri2014energy,lian2006multi}. Here we propose the use of supervised multi-view learning for classification of approximately-NST images that incorporate also structural information. We consider the texture and structure layers, decomposed from the original image, as two independent information sources. Features extracted from texture and structure are treated as two "views". Accordingly, two SVMs are trained on them independently. Lastly, the fusion of the two SVMs outputs is executed by training a shallow neural network, which optimizes the overall performance. Fusion of images via feed forward NNs can be also applied efficiently in the reconstruction of computed tomography \cite{boublil2015spatially}.     \\
	 \indent Experimental results show that our two-view method improves the overall performance compared with concatenating  the two views, or using the better one of them. This is demonstrated firstly in classification of images belonging to $6$ different substances, based on  the small subset of Kylbreg Texture Dataset \cite{Kylbreg2011c}. Wen then demonstrate the applicability of our proposed method, using breast ultrasound (BUS) images, based on the public (BUSIS) dataset \cite{xian2018benchmark}, where we classify the two classes of benign (B) and malignant (M) tumors. In \cite{lee2018intensity}, Lee et al. propose the use of intensity inhomogeneity correction and stacked denoising autoencoder (SADE) . Due to the small amount of data, the SADE yields the best performance compared with other deep learning methods. Here, we compare our method with the SADE and the four other methods mentioned in \cite{lee2018intensity}, and show that our proposed two-view achieves the best generalized performance.  
	 \section{Our Contribution}
	 The contributions of this work are:
	 \begin{enumerate}
	 	\item \textbf{Online Feature Extraction:} We extract meaningful features related to both structural and textural information. The extraction is performed online, without any learning process. These features facilitate interpretation of the nature of the data. In addition, the significance of this two-views approach stems from the fact that the structural and textural information complement each other.
	 	\item \textbf{Texture Modeling:} In contrast with existing supervised learning methods, we advance a classification technique that is based on texture modeling. We use the fBm model as prior of the textural layer (NST). Beyond that, we highlight the essence of the model by examining its two main characteristics: self-similarity and Gaussianity. 
	 	\item \textbf{Computational complexity:} compared with other deep learning methods, which are widely used for classification tasks, our method does not entail high computational complexity in terms of power and time. This is due to the fact that training the shallow fully connected NN, which has few  neurons, is not a heavy duty operation.
	 \end{enumerate}
	 \section{Background}	   
	\subsection{The fBm model}
	\label{sec:FBM}
\begin{figure}[t]
		\begin{minipage}[b]{0.31\linewidth}
			\centering
			\centerline{\includegraphics[width=4.8cm,height=3.9cm]{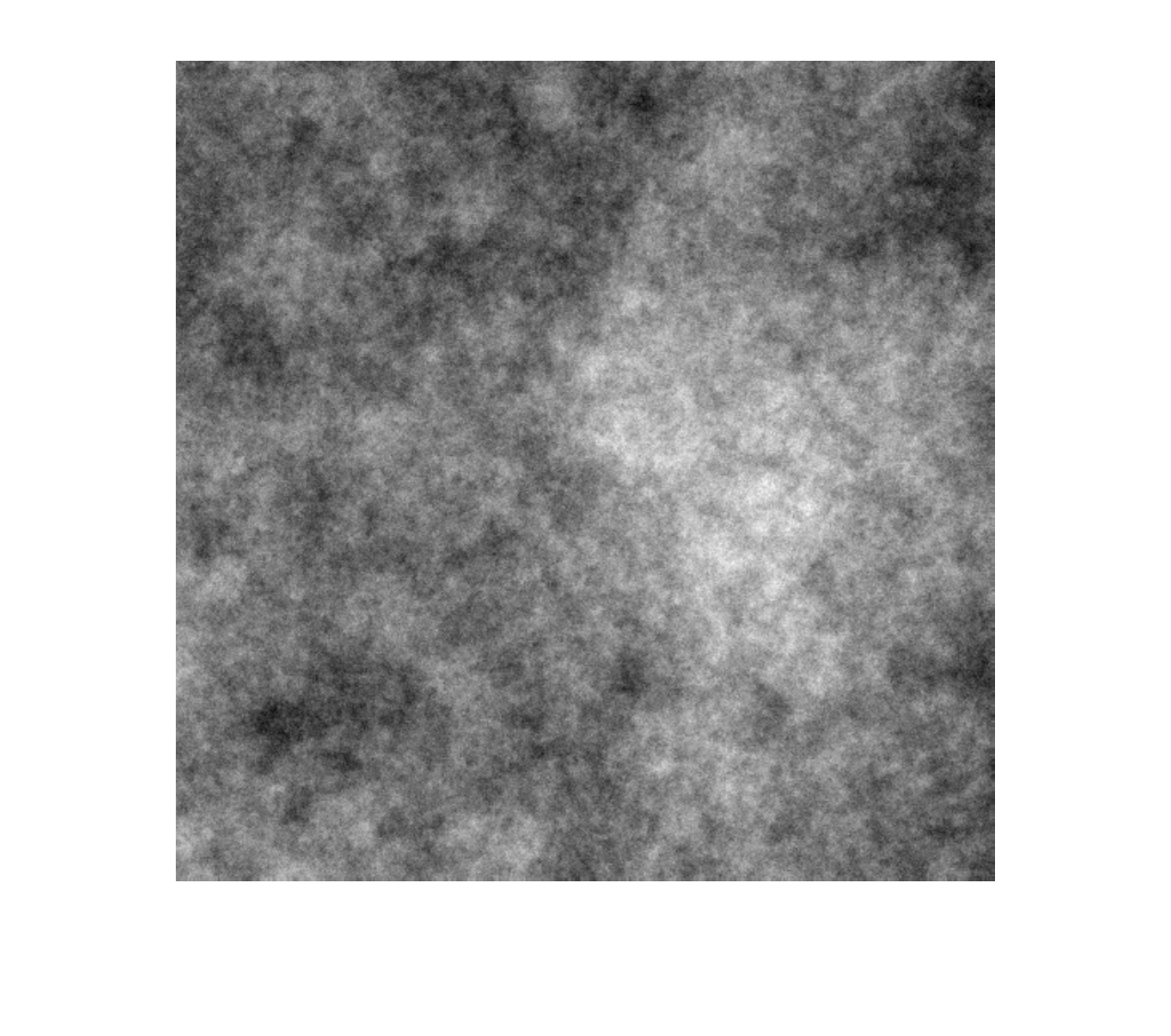}}
			\centerline{(a) \footnotesize  $\text{H}=0.2$}\medskip
		\end{minipage}
		\hspace{0.01mm}
		\begin{minipage}[b]{0.31\linewidth}
			\centering
			\centerline{\includegraphics[width=4.8cm,height=3.9cm]{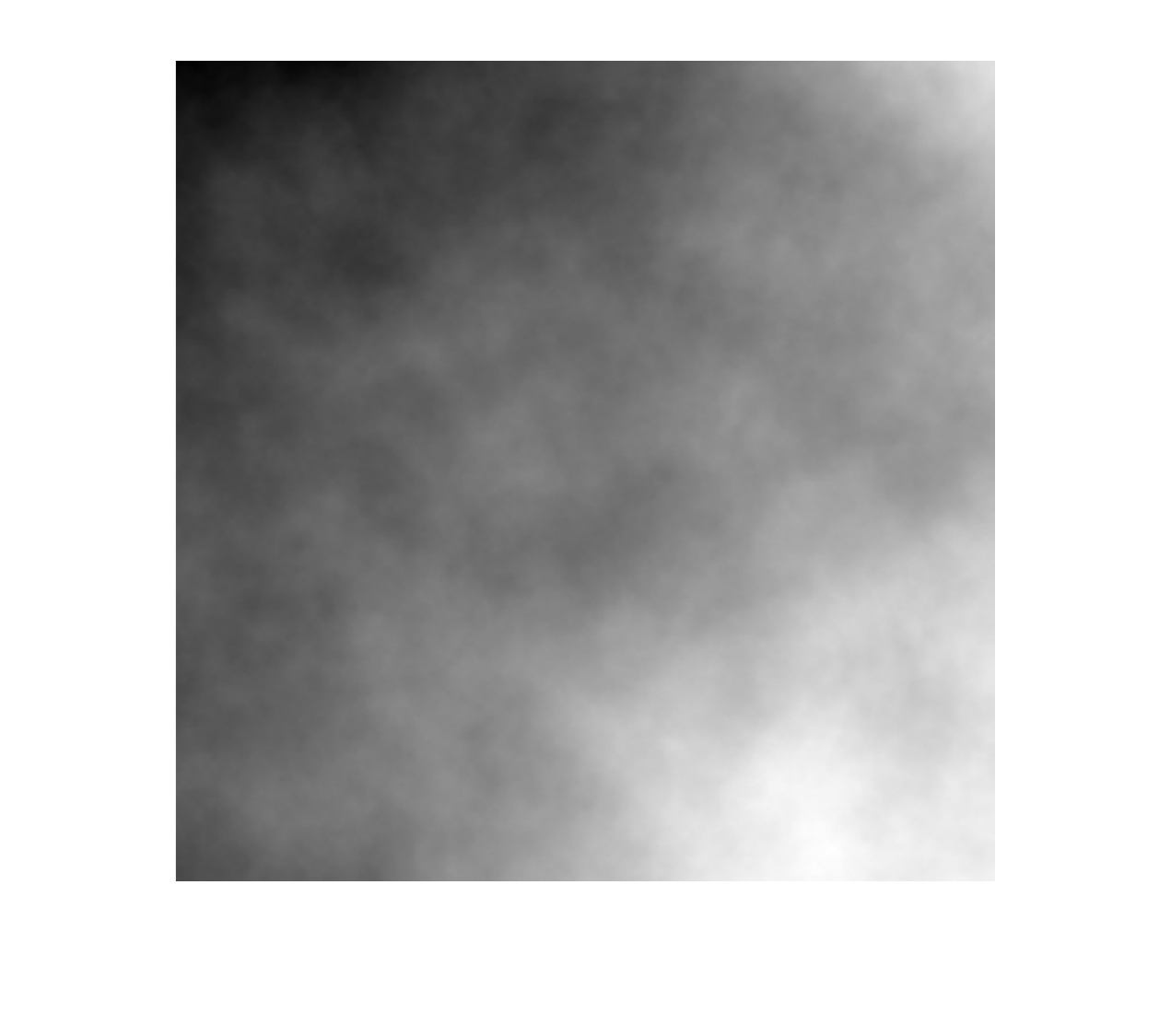}}
			\centerline{(b) \footnotesize  $\text{H}=0.8$ }\medskip
		\end{minipage}
		\hspace{1mm}
		\begin{minipage}[b]{0.3\linewidth}
			\centering
			\centerline{\includegraphics[width=4.8cm,height=3.9cm]{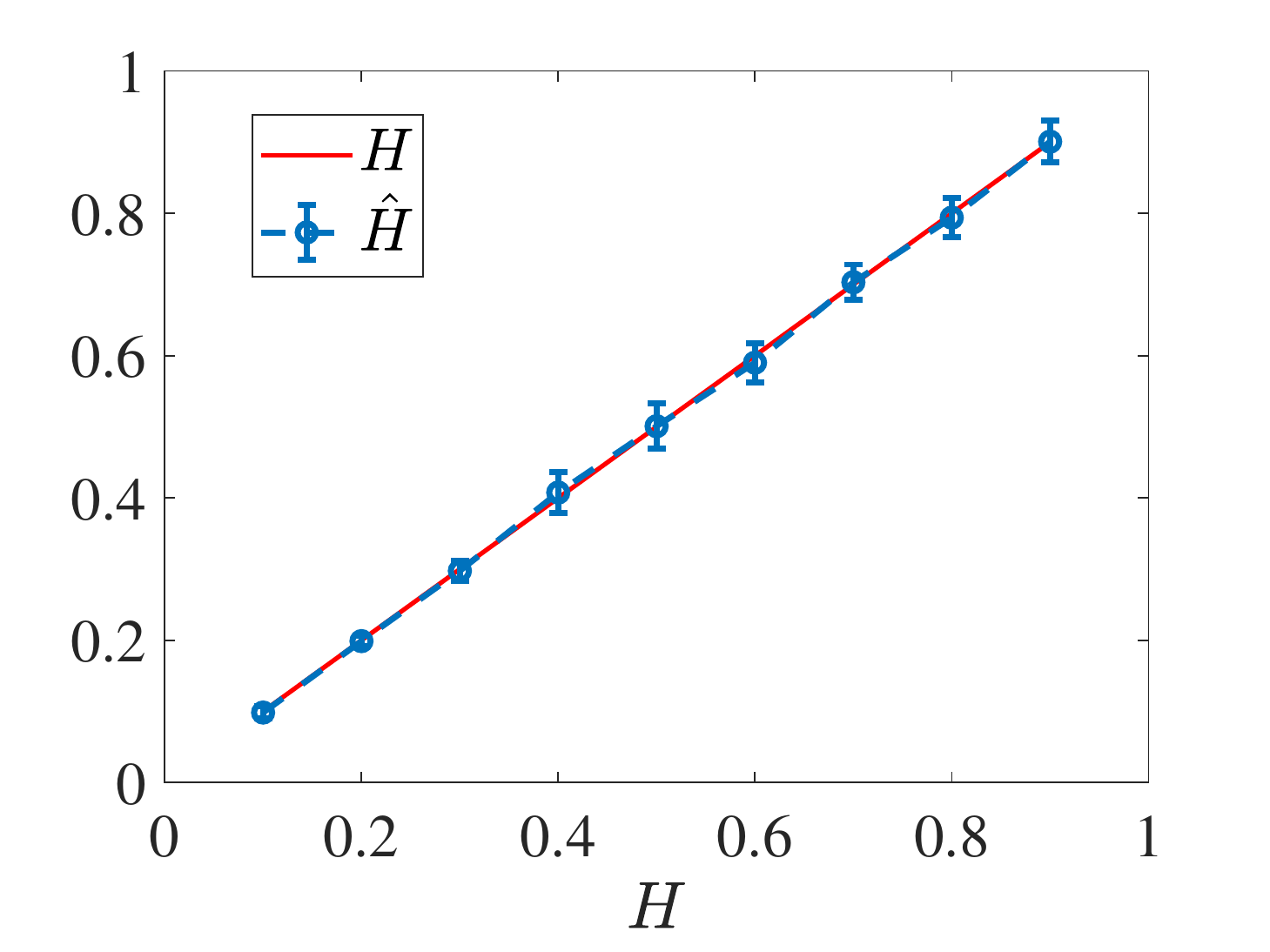}}
			\centerline{(c) \footnotesize $\text{H}$ Estimation}\medskip
		\end{minipage}
	\caption{Examples of synthetic fBm images, described by two values of $\text{H}$. (a) $\text{H}=0.2$ (b) $\text{H}=0.8$. (c) Estimated values of $\text{H}$ of synthetic fBm images results equal to real ground truth values with maximal estimator $\text{std}=0.028$ and maximal bias equal to $0.014$.}
		\label{fig:fBmSynth}
	\end{figure}

	FBm, introduced by Mandelbrot and Van Ness in 1968, is a continuous-time
	Gaussian process characterized by the following covariance: 
	\begin{equation}
	{\E}\left[B_{\text{H}}(t)B_{\text{H}}(s)\right]=\frac{\sigma_{\text{H}}^{2}}{2}(|t|^{2\text{H}}+|s|^{2\text{H}}-|t-s|^{2\text{H}})\label{eq:1},
	\end{equation}
	where 
	$\sigma_{\text{H}}^{2}=\sigma_{w}^{2}\cos(\pi\text{H})\Gamma(1-2\text{H})/2\pi\text{H}$
	where $\sigma_{w}^{2}$ is a known variance and $\text{H}\in(0,1)$
	is the so-called Hurst parameter or Hurst exponent \cite{Mandel}. This
	parameter determines the smoothness of the motion; higher values leading
	to a smoother motion. To highlight the meaning of H insofar as texture appearance is concerned, we present in Fig. \ref{fig:fBmSynth} an example of two synthetic fBm images with different H parameters. The first order increment
	of the process, $G_{\text{H}}(t)=B_{\text{H}}(t+1)-B_{\text{H}}(t)$
	is known as the fractional Gaussian noise (fGn). Since the fBm process is non-stationary, it is easier to study its increments, the fGn, which are stationary. The stationarity of the fGn lends itself to simple analysis and synthesis of images. This property will be exploited in the sequel analysis of NST images and Hurst parameter estimation.\\
	\indent The 2D generalization of the fBm process, called Lévy fractional Brownian field, is statistically isotropic \cite{FBM2002}. It has the following auto-correlation: 
	\begin{equation}
	{\E}\left[B_{\text{H}}(\textbf{x})B_{\text{H}}(\textbf{y})\right]=\frac{\sigma_{\text{H}}^{2}}{2}(||\textbf{x}||^{2\text{H}}+||\textbf{y}||^{2\text{H}}-||\textbf{x}-\textbf{y}||^{2\text{H}})\label{eq:2}
	\end{equation}
	where $\textbf{x}$ and $\textbf{y}$ are two points related to the 2D Euclidean space,
	$\textbf{x}=(x_{1},x_{2})$ and $\textbf{y}=(y_{1},y_{2})$, $||\cdotp||$ is the Euclidean
	norm operator and $\sigma_{\text{H}}^{2}$ is defined in \eqref{eq:1}. The variance of the increments: $B_{\text{H}}(x_{1},x_{2})-B_{\text{H}}(x_{1}-\Delta_{1},x_{2}-\Delta_{2})$, known as the structure function \footnote[1]{It is also known as variogram in geophysics} \cite{FBMstr,FBM2002}, is given by:
	\begin{equation} 
	\varphi(\Delta_{1},\Delta_{2})=f(\theta)(r)^{2\text{H}}\label{eq:3},
	\end{equation}
	where $r=\sqrt{\Delta_{1}^{2}+\Delta_{2}^{2}}$ and $f(\theta)=C$ in the case of isotropic fBm. This property can be exploited in estimation of the underlying H-parameter of the fBm process. According to \eqref{eq:3}, the empirical variance of the increments is a linear function of $r$ with slope of $2\text{H}$ on log-log scale. Therefore, the $\text{H}$ parameter can be estimated by using linear regression. Estimated $2\text{H}$ of synthetic fBm images are presented in Fig.\ref{fig:fBmSynth} (c).  
	
	\subsection{Texture and Structure Separation}
\begin{figure}[t]
 		\begin{minipage}[b]{0.22\linewidth}
 			\centering
 			\centerline{\includegraphics[width=3.5cm]{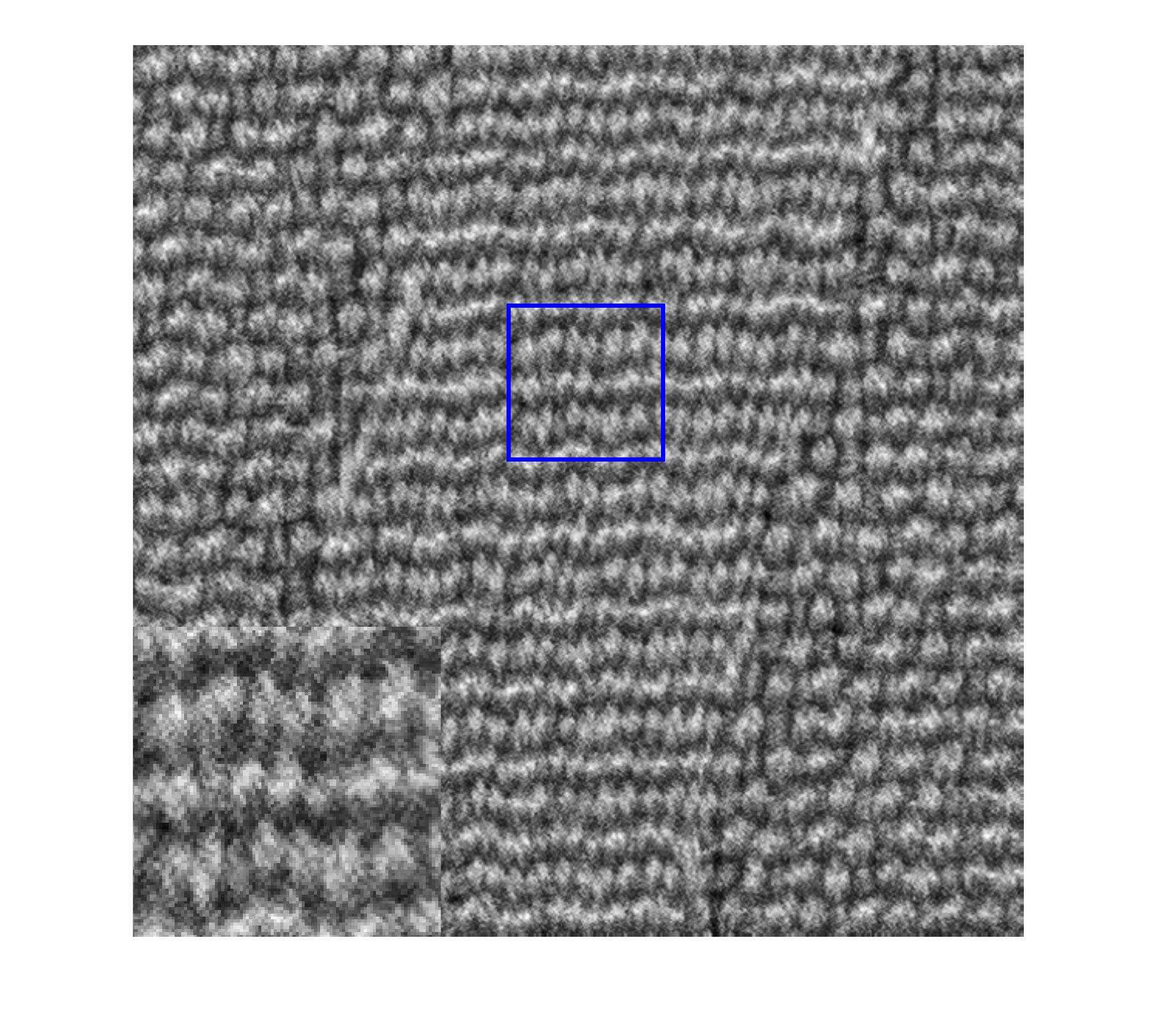}}
 			\centerline{(a) \footnotesize $I$}\medskip
 		\end{minipage}\hspace{0.01mm}
 		\begin{minipage}[b]{0.22\linewidth}
 			\centering
 			\centerline{\includegraphics[width=3.5cm]{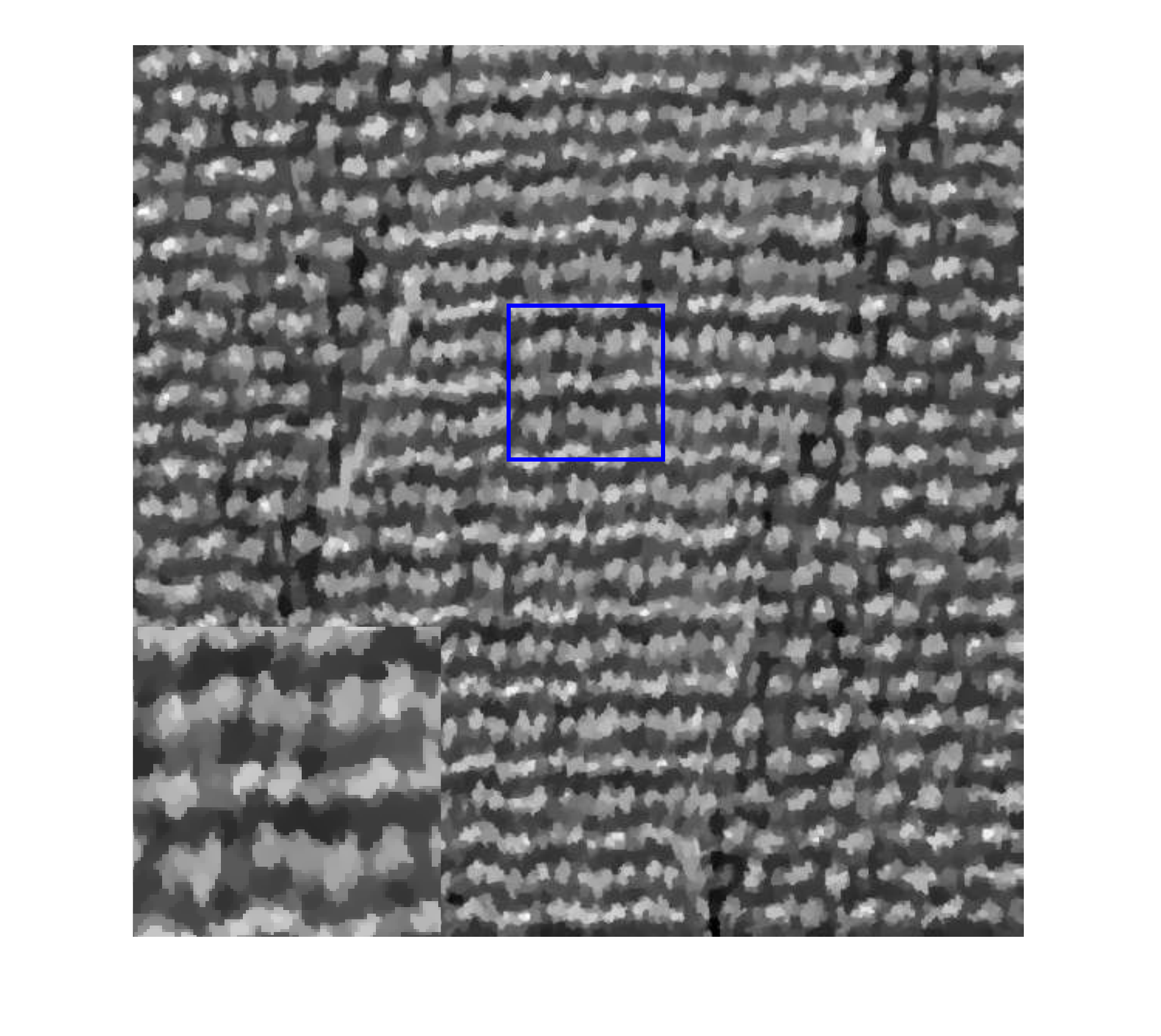}}
 			\centerline{(b) \footnotesize $S$}\medskip
 		\end{minipage}
 		\hspace{0.1mm}
 		\begin{minipage}[b]{0.22\linewidth}
 			\centering
 			\centerline{\includegraphics[width=3.5cm]{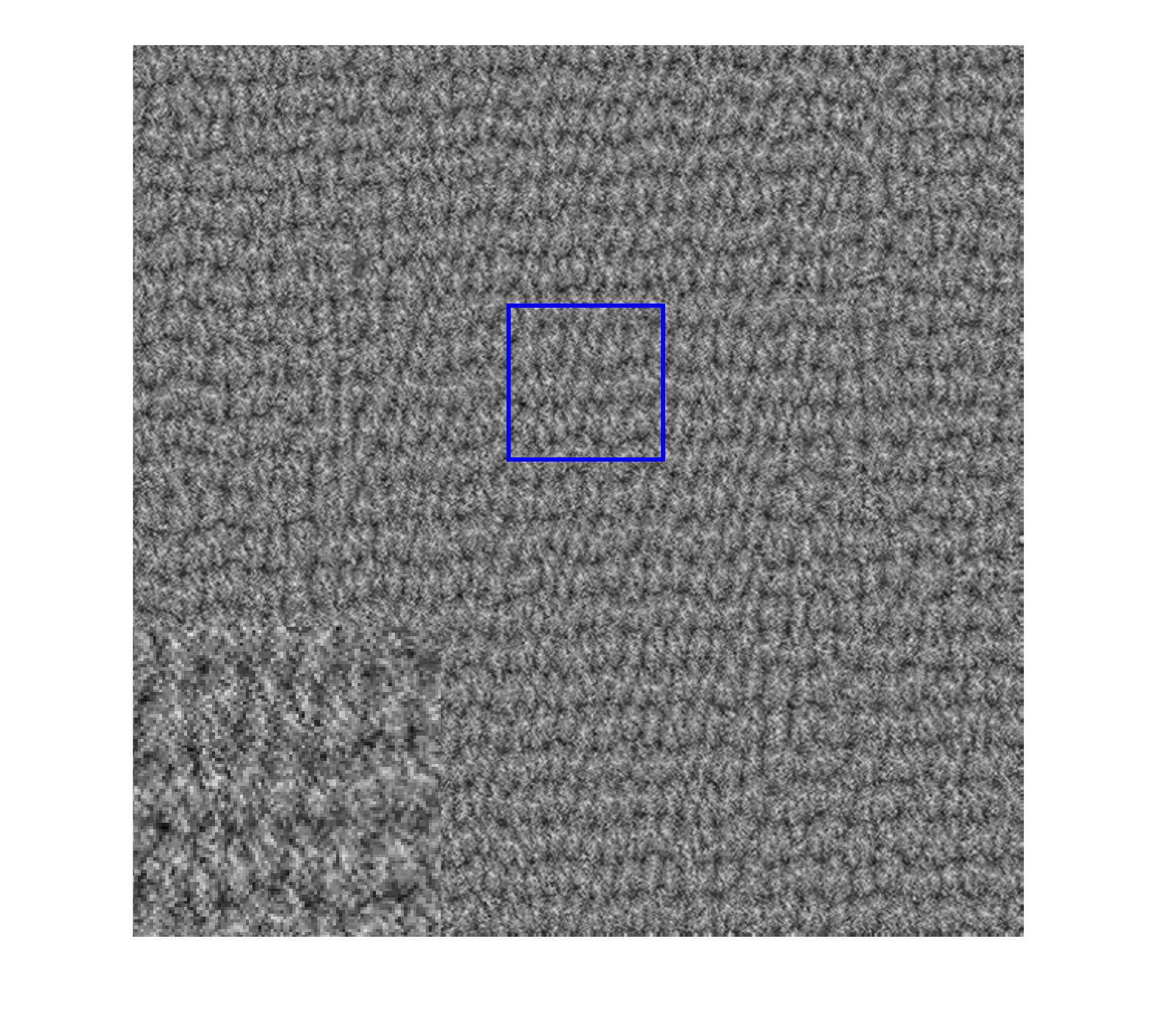}}
 			\centerline{(c) \footnotesize $T$ }\medskip
 		\end{minipage}
 		\hspace{0.1mm}
 		\begin{minipage}[b]{0.22\linewidth}
 			\centering
 			\centerline{\includegraphics[width=3.5cm,height=3.1cm]{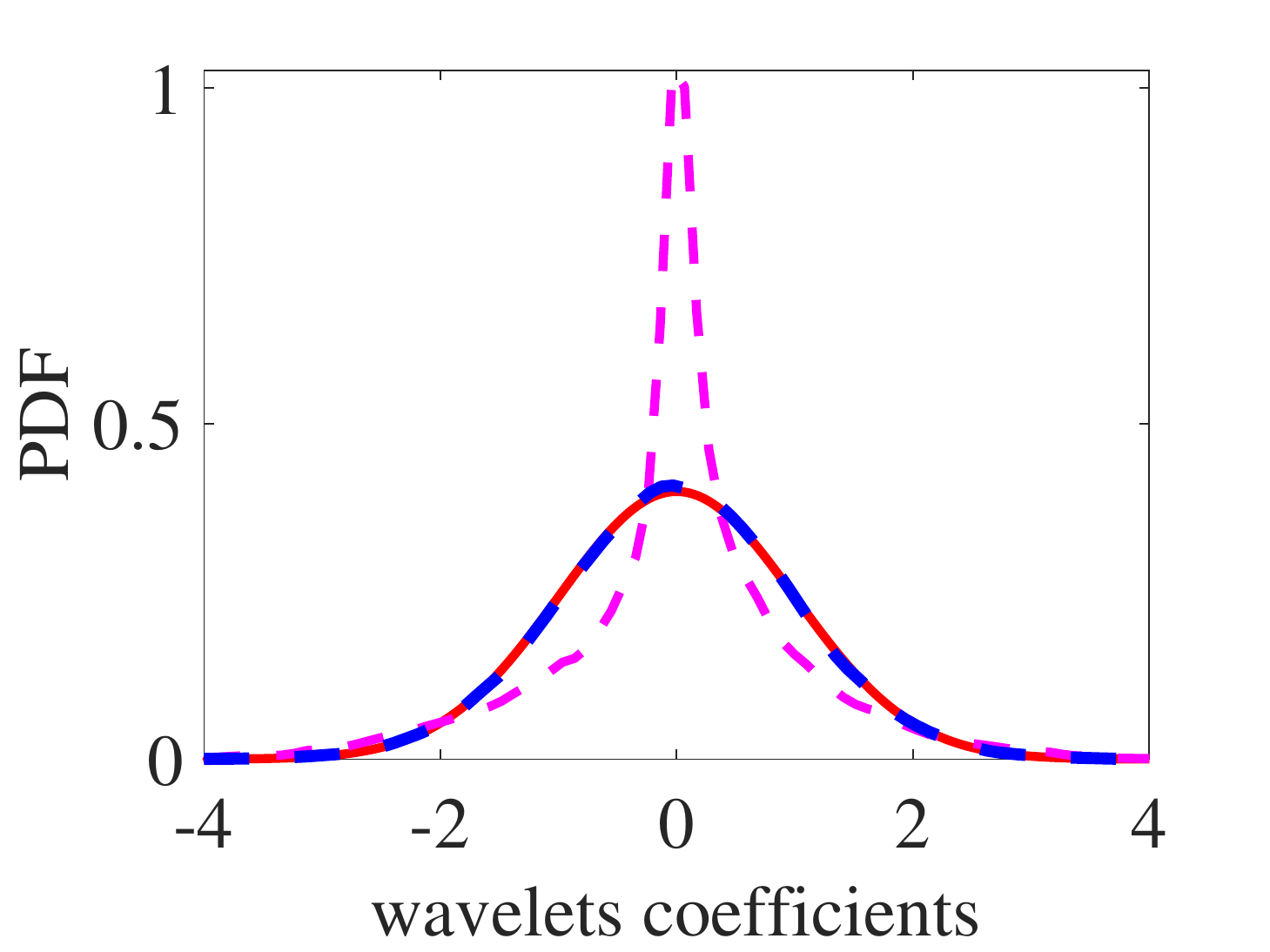}}
 			\centerline{(d) \footnotesize PDF }\medskip
 		\end{minipage}	
 		\\
 		\begin{minipage}[b]{0.22\linewidth}
 			\centering
 			\centerline{\includegraphics[width=3.5cm]{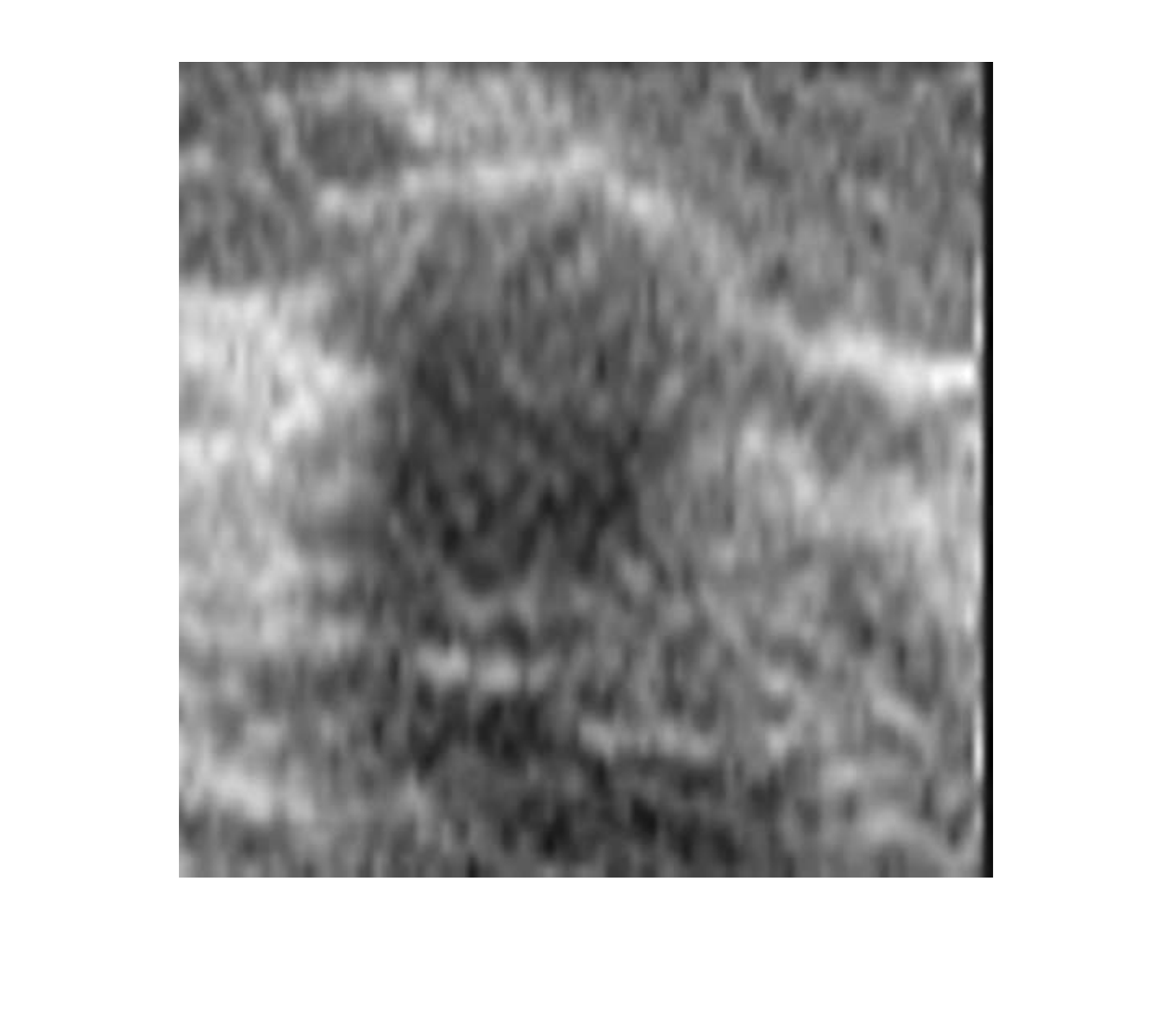}}
 			\centerline{(e) \footnotesize $I$ }\medskip
 		\end{minipage}
 		\begin{minipage}[b]{0.22\linewidth}
 			\centering
 			\centerline{\includegraphics[width=3.5cm]{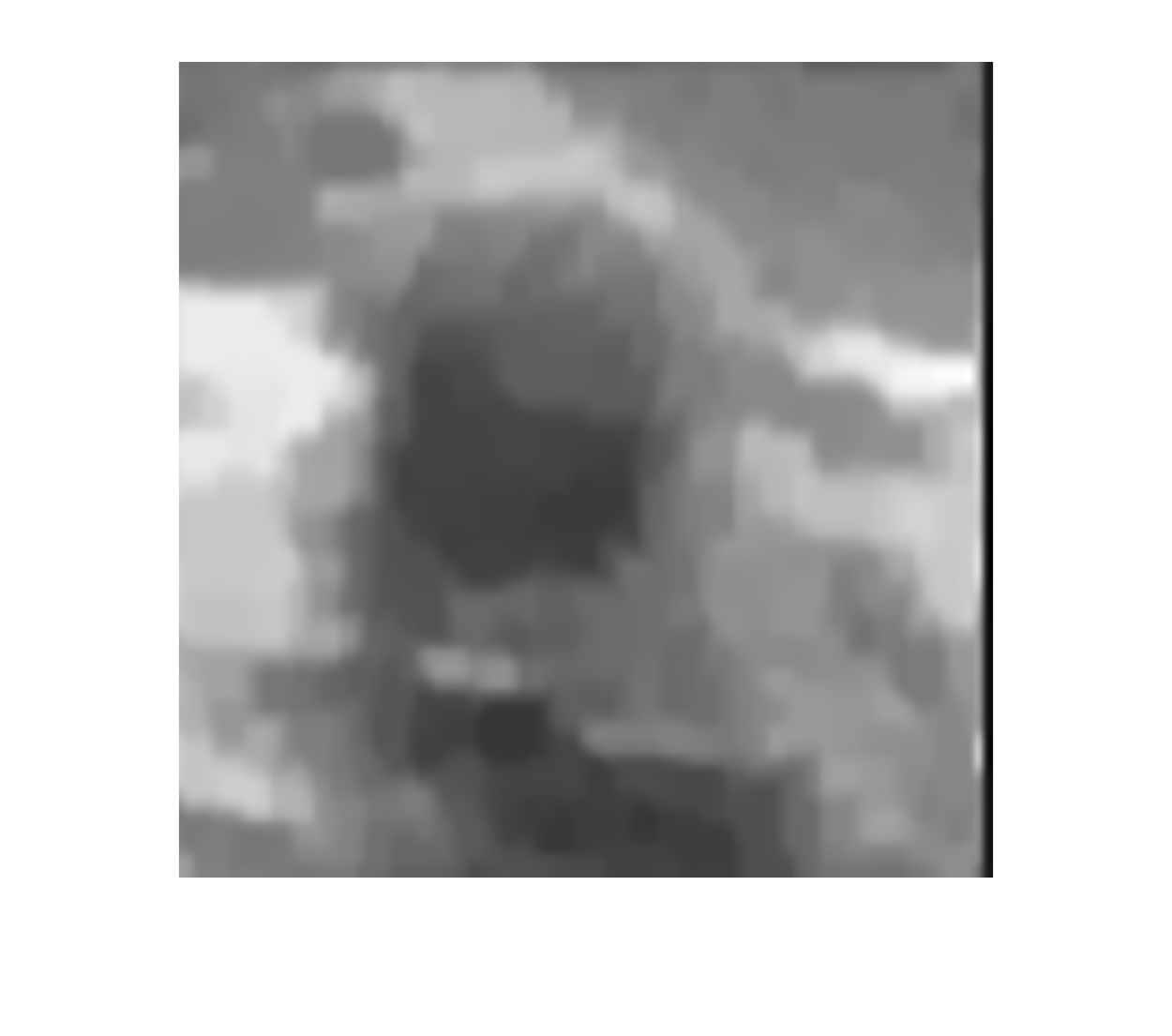}}
 			\centerline{(f) \footnotesize $S$ }\medskip
 		\end{minipage}
 		\hspace{0.1mm}
 		\begin{minipage}[b]{0.22\linewidth}
 			\centering
 			\centerline{\includegraphics[width=3.5cm]{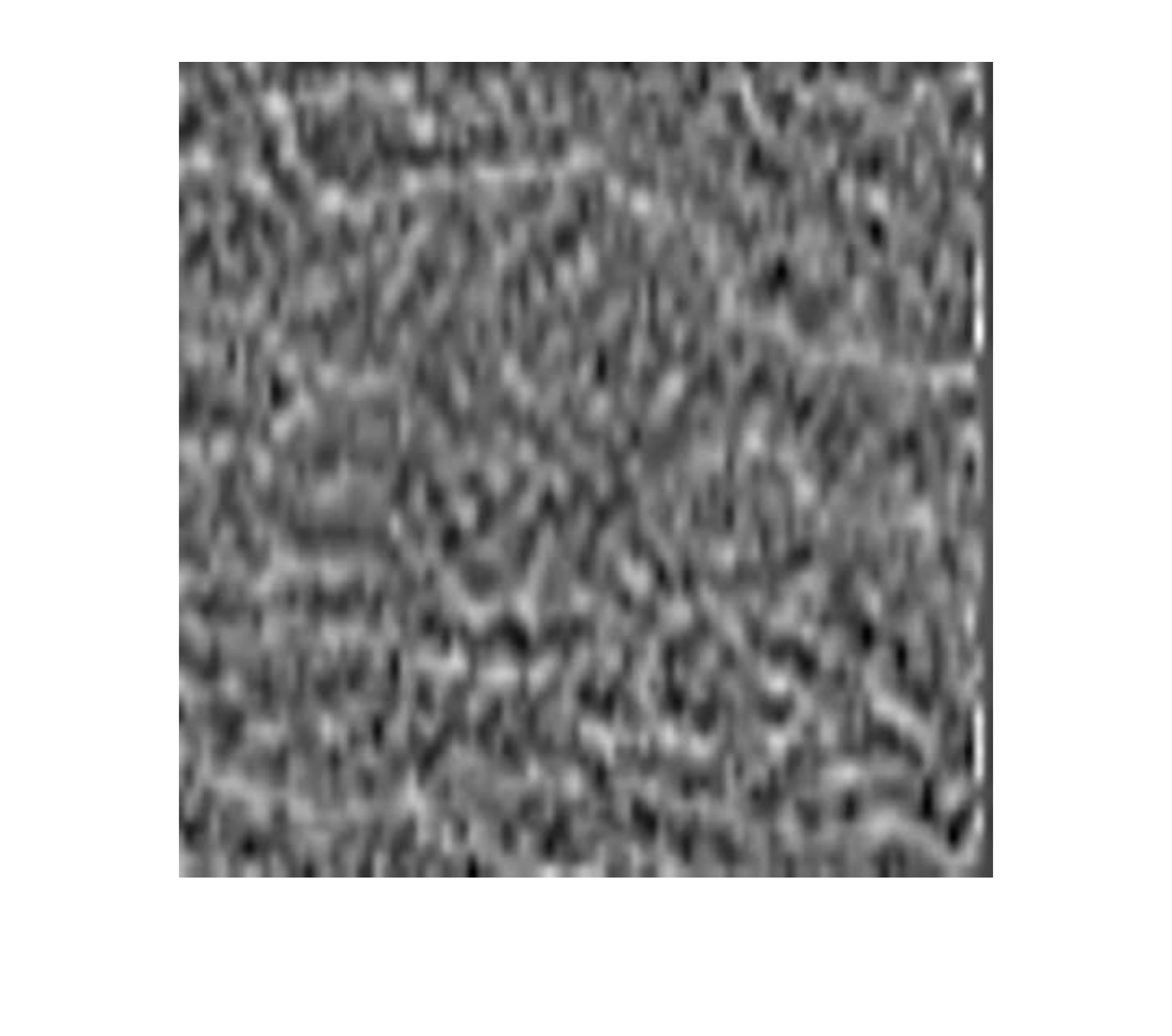}}
 			\centerline{(g) \footnotesize $T$ }\medskip
 		\end{minipage}
 		\hspace{0.1mm}
 		\begin{minipage}[b]{0.22\linewidth}
 			\centering
 			\centerline{\includegraphics[width=3.5cm,height=3.1cm]{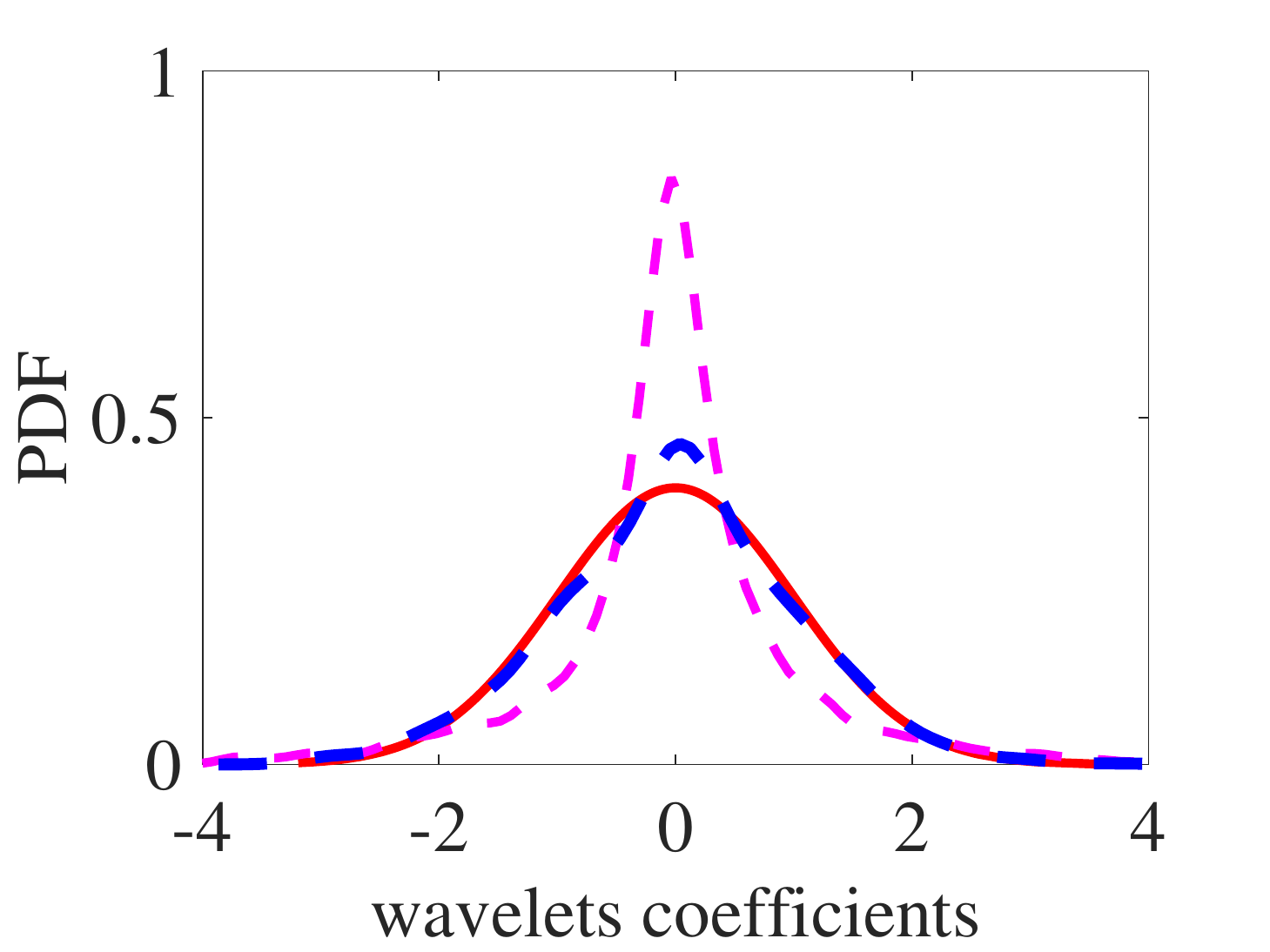}}
 			\centerline{(h) \footnotesize PDF }\medskip
 		\end{minipage}
 		\caption{The structured and textured image components obtained by RTV: (a,e) Examples of images downloaded from Kylbreg and BUSIS datasets, respectively. The ultrasound images is labeled as benign case. The corresponding structured components (b,f), $S$, exhibit edge or contour-like information, whereas the stochastic textured (c,g) components obey self-similarity and Gaussianity, as demonstrated in (d,h). The PDF of the wavelet coefficients of the textured layer (dashed blue) and the structured layer (dashed pink) are plotted with the best Gaussian fit PDF (solid red).} 
 		\label{fig:layers}
 	\end{figure}

 	Relative total variation (RTV) is a method that extracts structure from fully-textured images, which incorporate also patterned textures \cite{tsmoothing2012}. RTV solves an optimization system in which textural and structural edges are penalized differently \cite{tsmoothing2012}. Here we use the RTV to decompose the image into stochastic texture layer and a structured one. Let $I$ be the input image and $S$ denotes the RTV output image, which contains the extracted structural layer, while the residual image $T=I-S$ represents the textural layer. Examples of this decomposition applied on Kylbreg and BUSIS datasets, are illustrated in Fig.\ref{fig:layers}. As expected, the textural layer obeys Gaussianity (Fig.\ref{fig:layers}(d,h), dashed blue line), whereas, the structural component is characterized by high Kurtosis. This is obtained by calculating the empirical probability density function (PDF) of the wavelet coefficients of both layers.\\ 

	\section{The essence of self-similarity}\label{sec:ss}
	Self similarity exists in two senses: \textit{deterministic} and \textit{statistical}. The former is a geometrical property, which implies the existence of scale invariance in the image structure \cite{abry2015irregularities}. Geometrical fractals are typical examples of deterministic self-similar sets. However, the latter is less restrictive and it is generalization of the deterministic affinity to stochastic signals. Statistical self-similarity is manifested by the invariance of the probability law across scales, which implies that the random process shares in different scales the same statistics (PDF, moments). A typical example of self-similar process is the fractional Brownian motion (fBm) i.e: 
	$\forall\alpha>0\quad\mathrm{B}(\alpha t)\stackrel{d}{=}\alpha^{H}\mathrm{B}(t)$.
	 Given a self-similar process, its wavelets coefficients are also self-similar \cite{abry2015irregularities}. Let $t\in\mathbb{R}$ and the stochastic fBm process in 1D spatial/temporal domain be denoted by $B_{H}(t)$, then its wavelets coefficients are given as follows:
	\begin{equation}
	c_{j,k}=2^{j}\intop_{-\infty}^{\infty}B_{H}(t)\Psi(2^{j}t-k)\,dt \label{eq:4},
	\end{equation}
	where $\{ 2^{\frac{j}{2}}\Psi(2^{j}t-k)\:\forall j,k\in\mathbb{R}\} $ are the wavelets basics of $L^{2}(\mathbb{R})$ and $\Psi(t)$ is the mother wavelet function. The coefficients $\left\{ c_{j,k}\:\forall j,k\in\mathbb{R}\right\}$ are also Gaussian, as they are expressed as an integral of a Gaussian process
	$B_{H}(t)$. The following proposition expresses the scale-invariance property: 
	\begin{prop}\label{prop1}
		If \textup{$B_{H}(t)$} is a self-similar fBm process with Hurst
		exponent $H$, and $\{c_{j,k}\}$ are its wavelets coefficients as expressed
		in \eqref{eq:4}, then:
			\[
			\forall j,k\in\mathbb{R}\qquad c_{j,k}\stackrel{d}{=}2^{-H}c_{j-1,k}
			\]	
	\end{prop}
		\indent This proposition can be directly proven by showing the invariance of the two-first moments along scales or levels, i.e. $j$, thanks to the Gaussianity of $c_{j,k}$\footnote[2]{The complete proof of Proposition.\ref{prop1} and Proposition.\ref{prop2} \label{ft1} addressed in section \ref{mathproofs}.}. The aforementioned proposition implies that the statistics of the wavelets coefficients are preserved along scales (i.e the probability density, mean, variance and higher moments). The importance of this fact was mentioned in \cite{abry2015irregularities} and it is exploited elsewhere in Hurst exponent estimation for both 2D and 1D signals \cite{veitch1999wavelet}. However, here we use this proposition in a different context, to asses the self-similarity of the NST by calculating the distance between the PDFs of wavelet coefficients at different levels. The distance between the PDFs of the self-similar wavelet coefficients should be zero. We therefore calculate the distance between the PDFs of the NST wavelets coefficients in two different levels \footnote[3]{For simplicity, here we consider the marginal, not the joint, PDFs of the coefficients.}. Given that the underlying process is fBm, and under the assumption of zero-mean coefficients, the Kullback–Leibler (KL) divergence between two levels is given by \cite{kullback1951information}:
		\begin{equation}
		D_{KL}(p_{1}||p_{2})=\log\left(\frac{\sigma_{2}}{\sigma_{1}}\right)+\frac{\sigma_{1}^{2}}{2\sigma_{2}^{2}}-\frac{1}{2}\label{eq:5},
		\end{equation} 
		where $p_{1}$ and $p_{2}$ are the PDFs of the coefficients $c_{j,k}$ and $2^{-H}c_{j-1,k}$ of the multiresolution wavelet representation, respectively.\\
		\indent Practically, the variance of the wavelet coefficients is estimated using the Maximum likelihood (ML). The ML estimator is consistent, i.e. it convergences to the real parameter ($\sigma$) when the number of samples is large enough. In the case of NST, due to scale-invariance, the ML variance estimator at different wavelets levels converges to the same variance. As a sequence of the consistency and scale-invariance of the NST wavelets, we arrive at the following proposition\textsuperscript{\ref{ft1}}:        
		\begin{prop}\label{prop2}
		Let \textup{$\hat{\sigma}_{1}$and $\hat{\sigma}_{2}$} be the ML
		estimators of the wavelets coefficients $c_{j,k}$ and $2^{-H}c_{j-1,k}$ of the self-similar fBm, respectively, then $\exists$ $c>0$ s.t 
		\[
		D_{KL}(p_{1}||p_{2})<c
		\]
		\end{prop}
	  		
		\indent In our experiments, we adopt the use of KL divergence and other norm-induced distances to assess the self-similarity, or scale-invariance, of the NST. Alternatively, the latter can be quantified by measuring the mutual information exists between different image resolutions or blocks \cite{khawaled2019self,alexander2007simple}.    
		\subsection{Experiments with Different Distance Measures}
\begin{figure}[t]
		\begin{minipage}[b]{0.41\linewidth}
			\centering
			{\includegraphics[width=5.8cm]{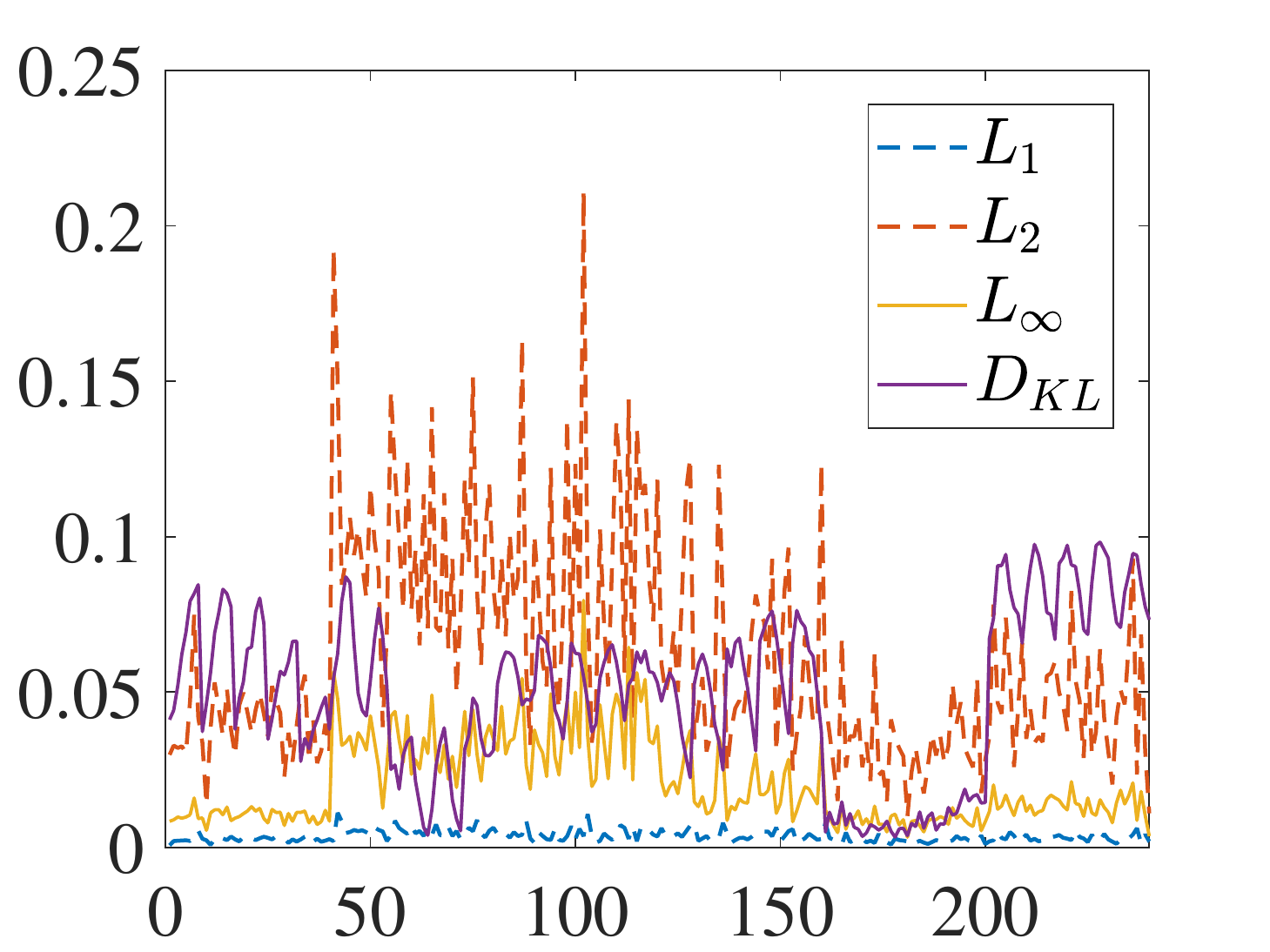}}
			{(a)}\medskip
		\end{minipage}
		\begin{minipage}[b]{0.41\linewidth}
			\centering
			{\includegraphics[width=5.8cm]{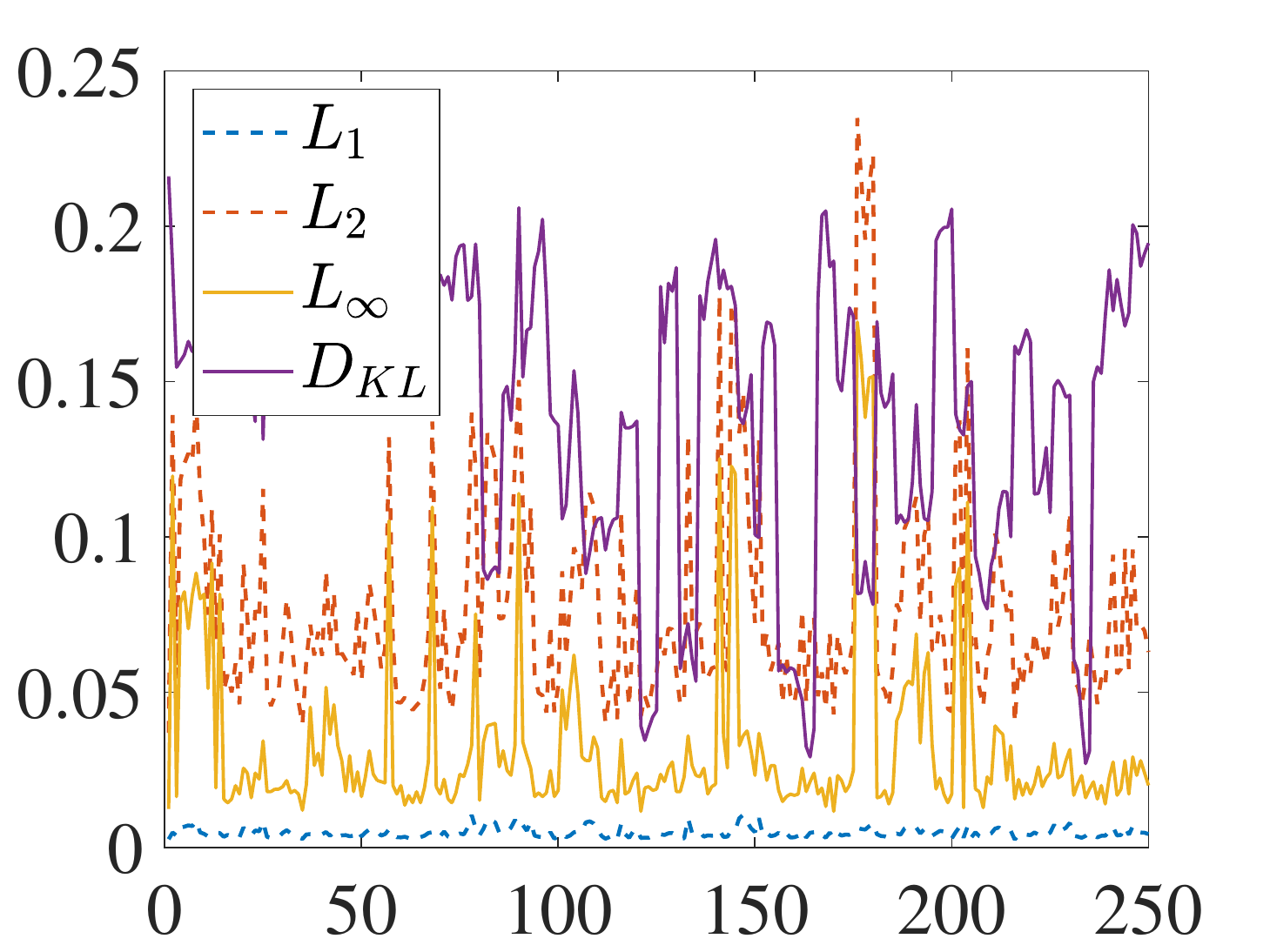}}
			{(b)}\medskip
		\end{minipage}
		\caption{Assessing self-similarity by calculating the distance between the wavelets' PDFs. Shown are the distances between the PDFs of level $1$ and level $3$ for Kylbreg (a) and BUSIS (b) datasets. It may be concluded that $L_{1}$-induced metric yields the most robust and smallest distance.}\label{fig:SSdis}
	\end{figure}

		 \indent After having separated texture and structure parts, obtained by performing RTV on images from both datasets, we proceed to show that the main characteristic of the fBm model, namely the self-similarity, exists in the NST component. We quantify the self-similarity in the wavelet domain, as described earlier in section \ref{sec:FBM}, by calculating the distance between the PDFs of wavelets coefficients in two different levels. In practice, estimation of the PDF can be easily accomplished by means of the first two moments, using the ML estimators, as a consequence of Gaussianity of the wavelets coefficients.\\
		\indent In our experiments, we calculate the first $3$ levels of the Haar wavelets of the textured layer, for both BUSIS and kylbreg datasets. We focus only on levels $1$ and $2$, where we estimate their principle moments and PDFs. Then, the KL divergence between their PDFs is calculated according to \eqref{eq:5}. In addition, for further verification of the essence of self-similarity, we calculate other norm-induced distances between the PDFs of levels $1$ and $3$, including the $L_{2}$, $L_{\infty}$ and $L_{1}$, which also is known as the TV distance. Results of the distances calculation on both datasets are presented in Fig.\ref{fig:SSdis}.   
	\section{The Two-View Texture-Structure Method}
\begin{figure}[t]
		\includegraphics[width=12.5cm,height=4.9cm]{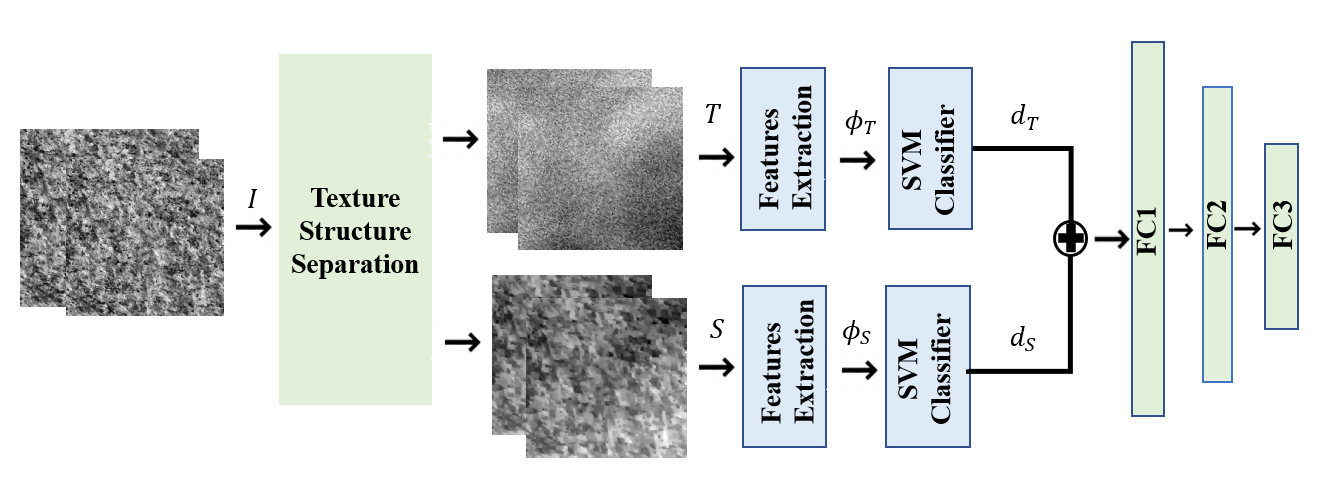}
		\centering
		\caption{The proposed two-view classification system.}
		\label{fig:system}  
	\end{figure}

	This section introduces our proposed classification method. Firstly, we decomposed the images into two separated layers: texture and structure. Then we extract the two-view features: structure and texture related features. Afterwards, two independent SVMs are trained on these features separately, as described in Fig.\ref{fig:system}. Afterwards, the distance between the image example and the hyper-plane of each SVM is used as an input to shallow neural network (NN) constructed from 3 fully connected layers. It should be noted that training the network is done on examples that are not included in the training set of the two SVMs, that's how we can fuse two views efficiently. Details about the training procedure of both the SVMs and the NN are addressed in the following subsections.
	\subsection{Features Extraction}\label{sec:feat}
	\subsubsection{Textural Features}
\begin{figure}[t]
			\begin{minipage}[b]{0.31\linewidth}
				\centering
				{\includegraphics[width=4.9cm]{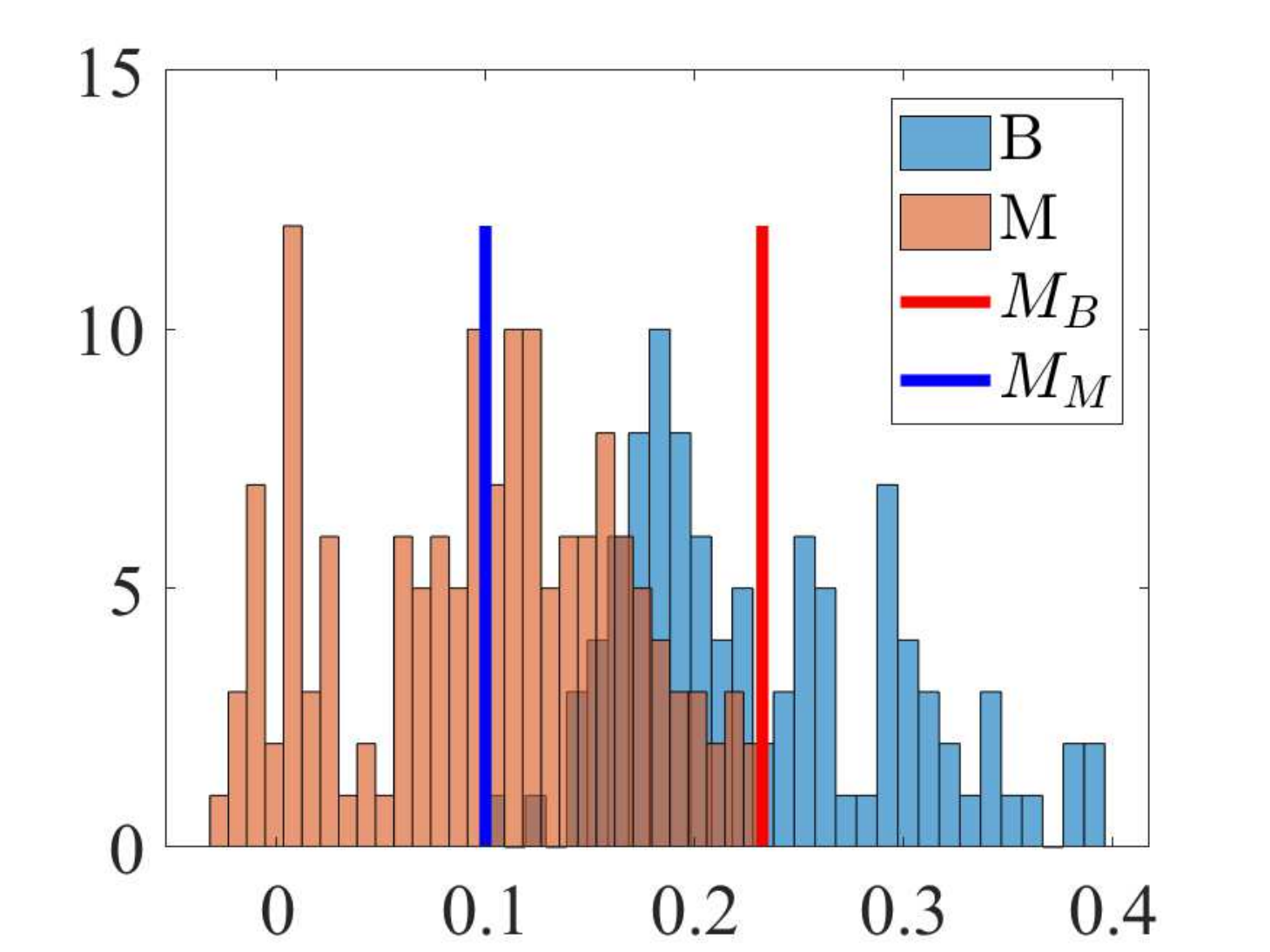}}
				{(a)}\medskip
			\end{minipage}\hspace{0.01mm}
			\begin{minipage}[b]{0.31\linewidth}
				\centering
				{\includegraphics[width=4.9cm]{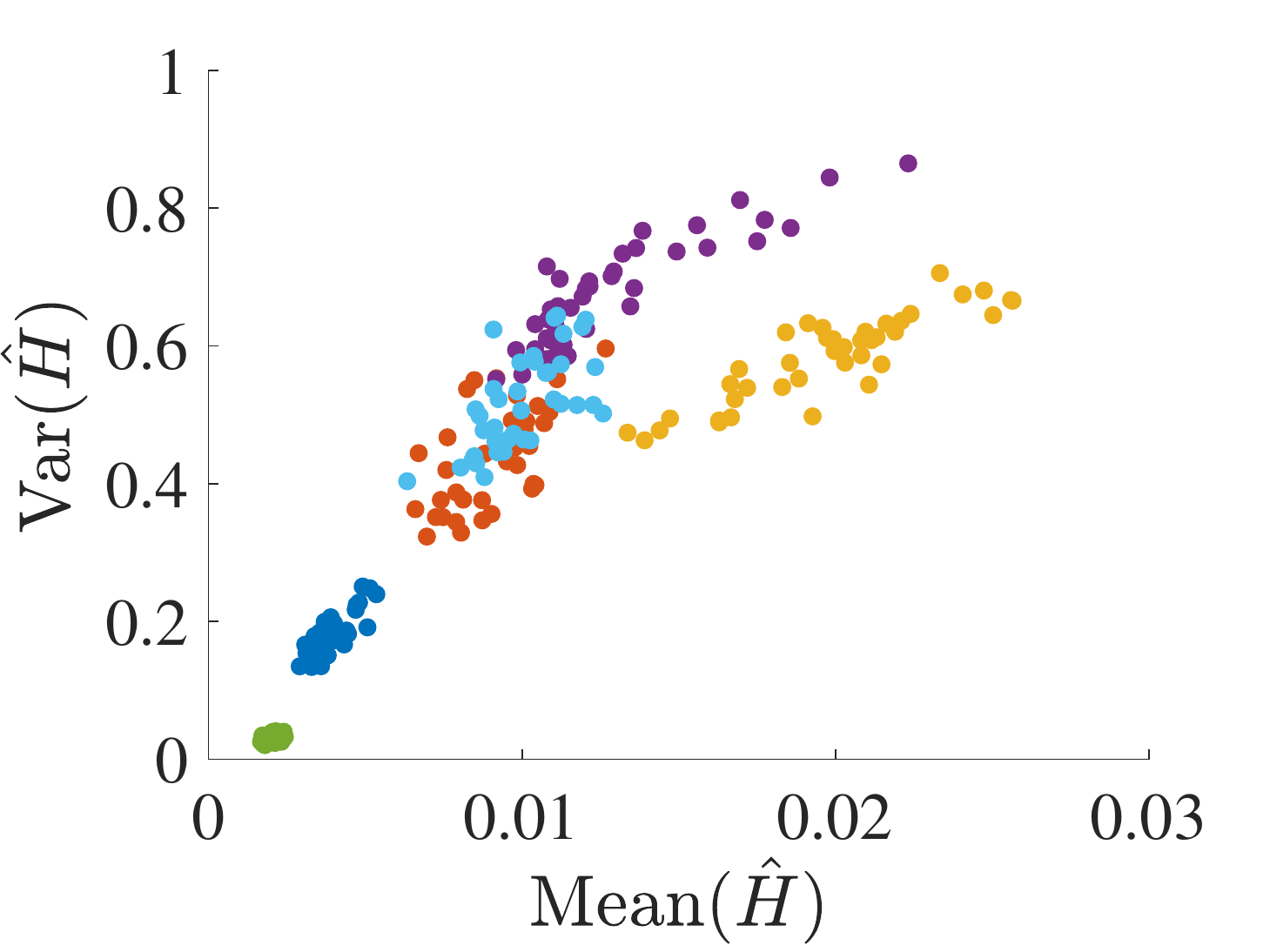}}
				{(b)}\medskip
			\end{minipage}
			\hspace{0.1mm}
			\begin{minipage}[b]{0.31\linewidth}
				\centering
				{\includegraphics[width=4.9cm]{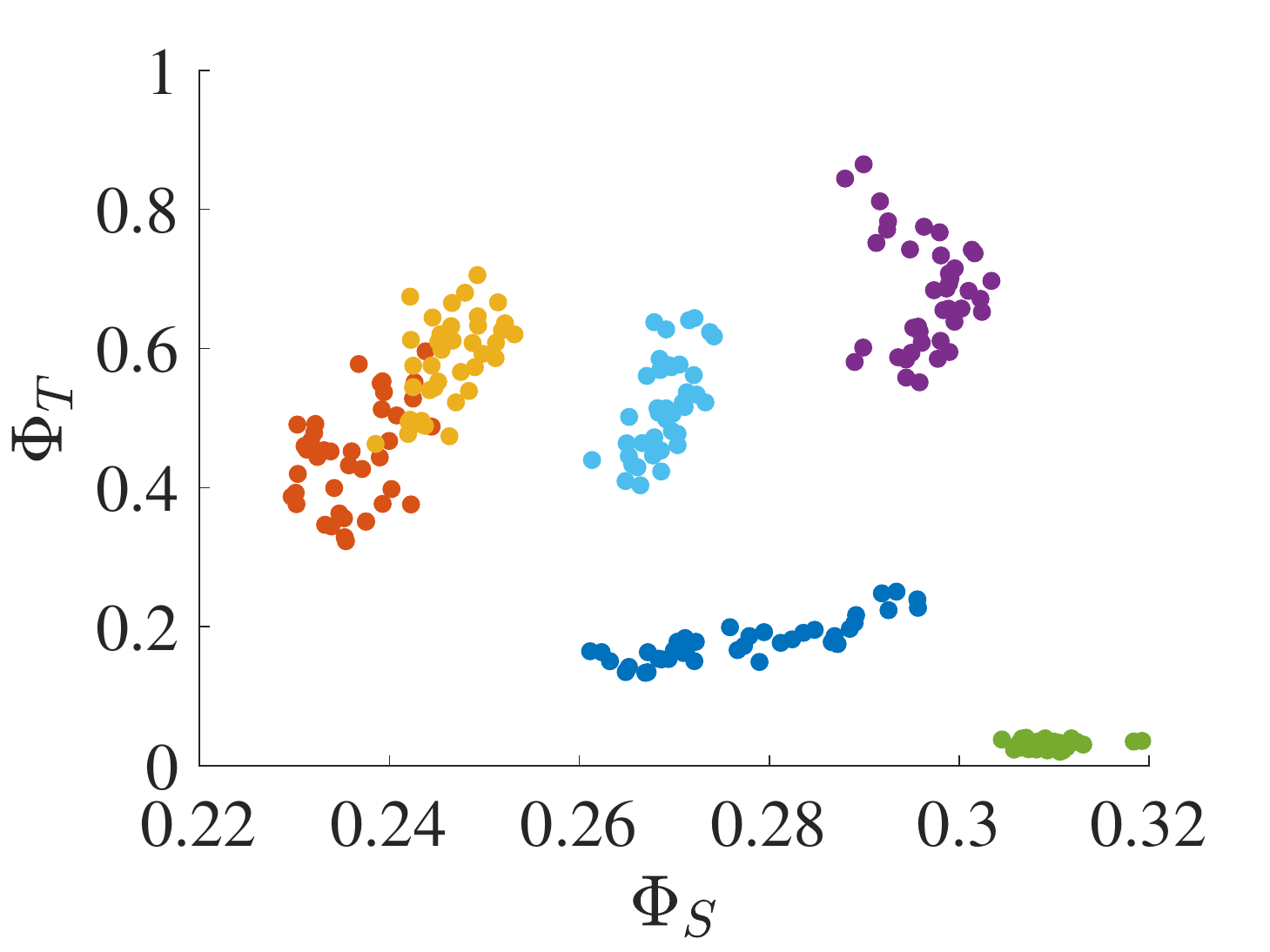}}
				{(c)}\medskip
			\end{minipage}
			\caption{Textural and structural features. (a) The distribution of the mean Hurst parameter over patches of benign (B) and Malignant (M) data, with the annotated mean value represented for the corresponding classes $M_{B}$ and $M_{M}$. (b) Scatter plot of mean and variance of estimated H over patches for the kylperg dataset, different colors corresponds to different classes. (c) Scatter plot of mean H over patches and mean phase congurrency for the kylperg dataset. Combining the structural ($\Phi_{S}$) and textural ($\Phi_{T}$) information improves the distinctions between different classes.} \label{fig:TF}
		\end{figure}

		The 2D fBm is a suitable model for the separated NST layer, in that the separated textures satisfy the Gaussianity (as illustrated in Fig\ref{fig:layers}) and self-similarity, which we assessed in the wavelet domain for examples from both datasets (further discussed in Section \ref{sec:ss}). In view of the fitness of these characteristic of fBm to such NST images, we proceed to estimate the Hurst parameter and use it as a main feature. As already mentioned in Section \ref{sec:FBM}, the estimated Hurst parameter characterizes the roughness of the texture. We therefore expect that this parameter will enhance the distinction between different patterns of NSTs, as is called for the case of classification of benign and Malignant tissues.
		In our experiments, we divide the textural image into non-overlapping patches of size $32\times32$ and estimate the Hurst parameter as outlined in Section \ref{sec:FBM}. Figure.\ref{fig:TF}(a) shows the distribution of this parameter across the two populations of M and B. Figure.\ref{fig:TF}(b) presents the mean and variance of the estimated H, computed over patches, extracted from the textural layer of the Kylbreg dataset. Note that this parameter contributes significantly to distinction between different textures. Nevertheless, it is still necessary to combine additional features to obtain better separation between the corresponding classes. This is the case in both examples. An additional feature is related to the complementary structural information.	 
		\subsubsection{Structural Features}
\begin{figure}[!b]
			\begin{minipage}[b]{0.31\linewidth}
				\centering
				{\includegraphics[width=4cm]{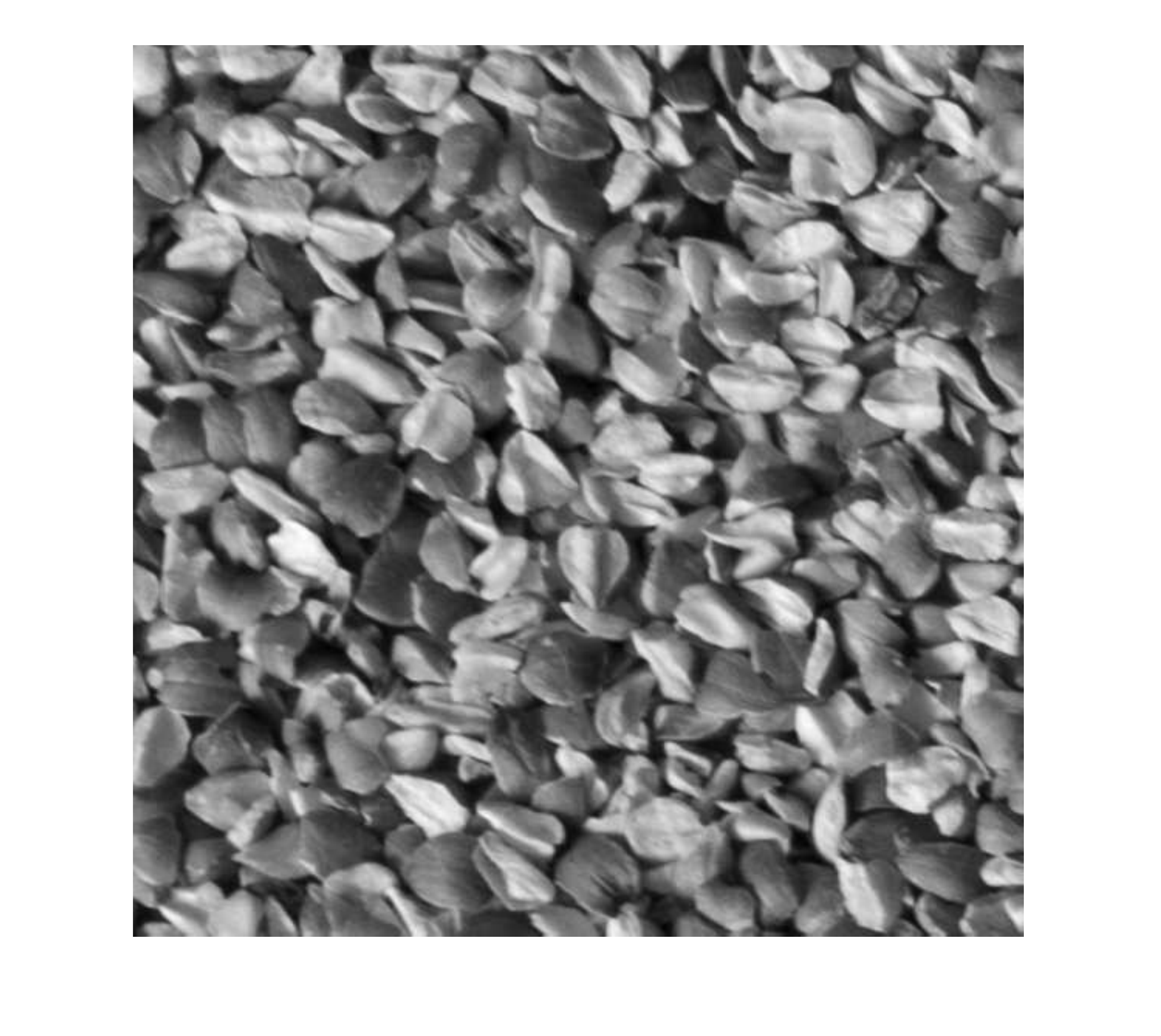}}
				{(a)}\medskip
			\end{minipage}\hspace{0.01mm}
			\begin{minipage}[b]{0.31\linewidth}
				\centering
				{\includegraphics[width=4cm]{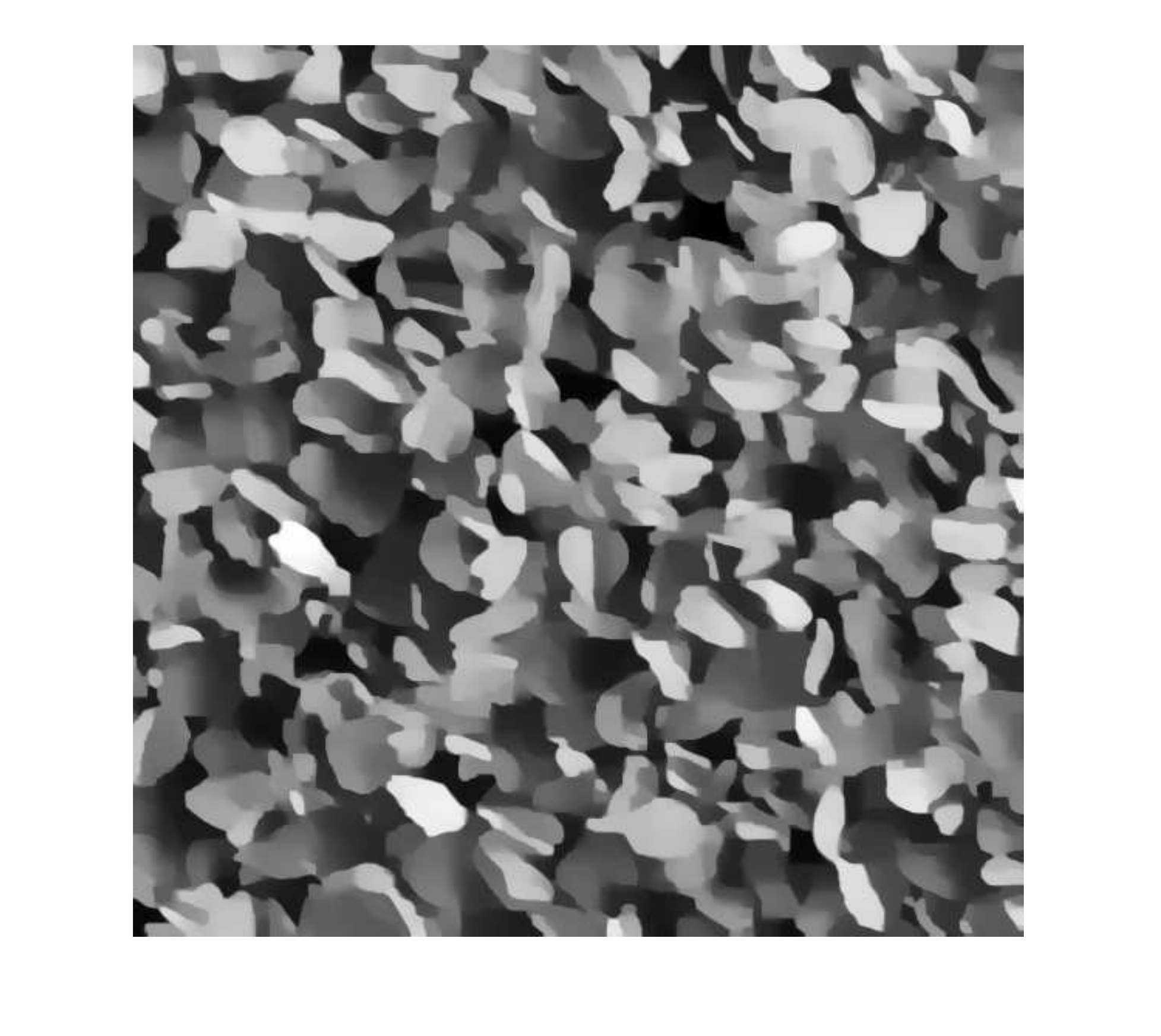}}
				{(b)}\medskip
			\end{minipage}
			\hspace{0.1mm}
			\begin{minipage}[b]{0.31\linewidth}
				\centering
				{\includegraphics[width=4cm]{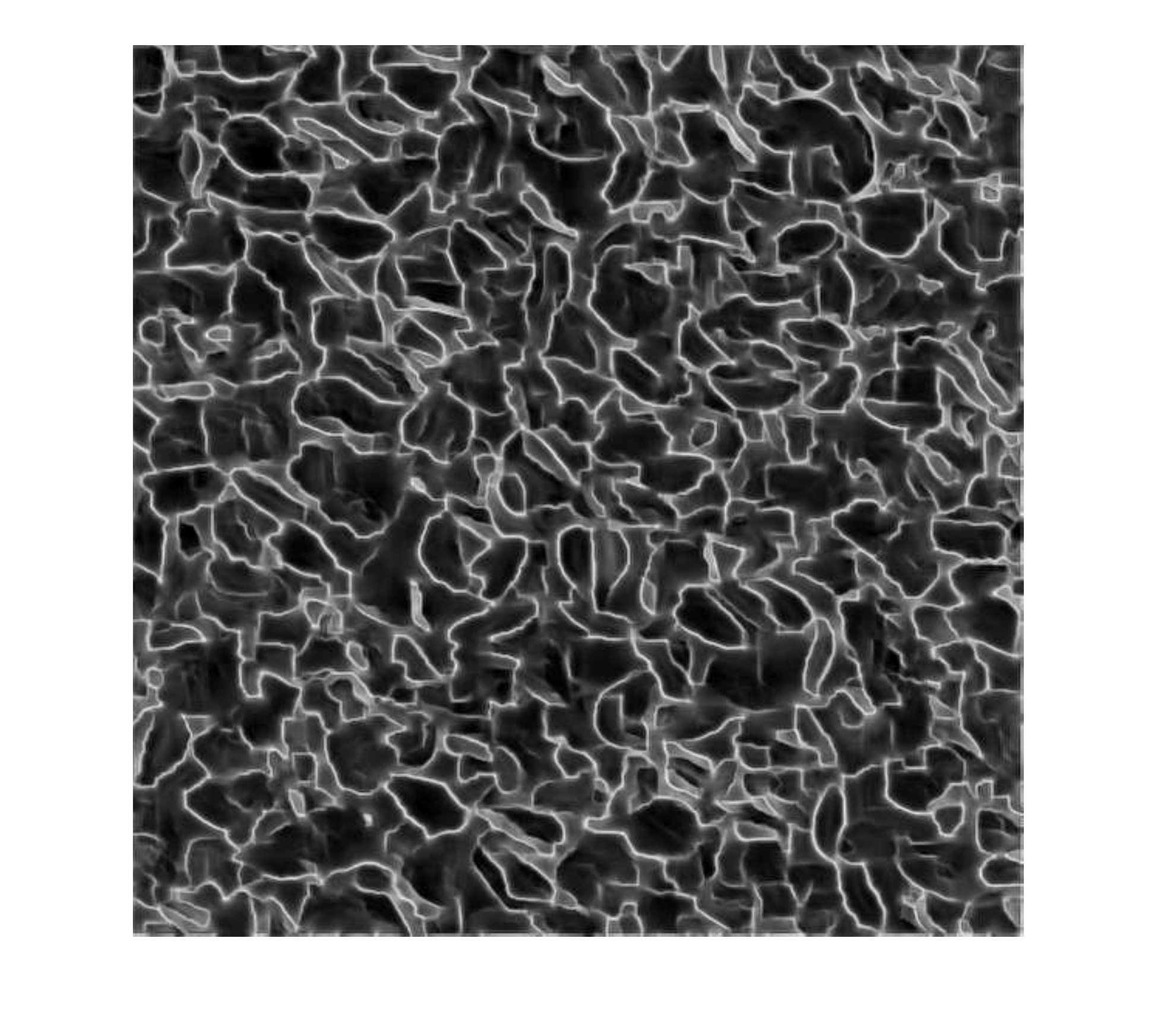}}
				{(c)}\medskip
			\end{minipage}\\
			\begin{minipage}[b]{0.31\linewidth}
				\centering
				{\includegraphics[width=4cm]{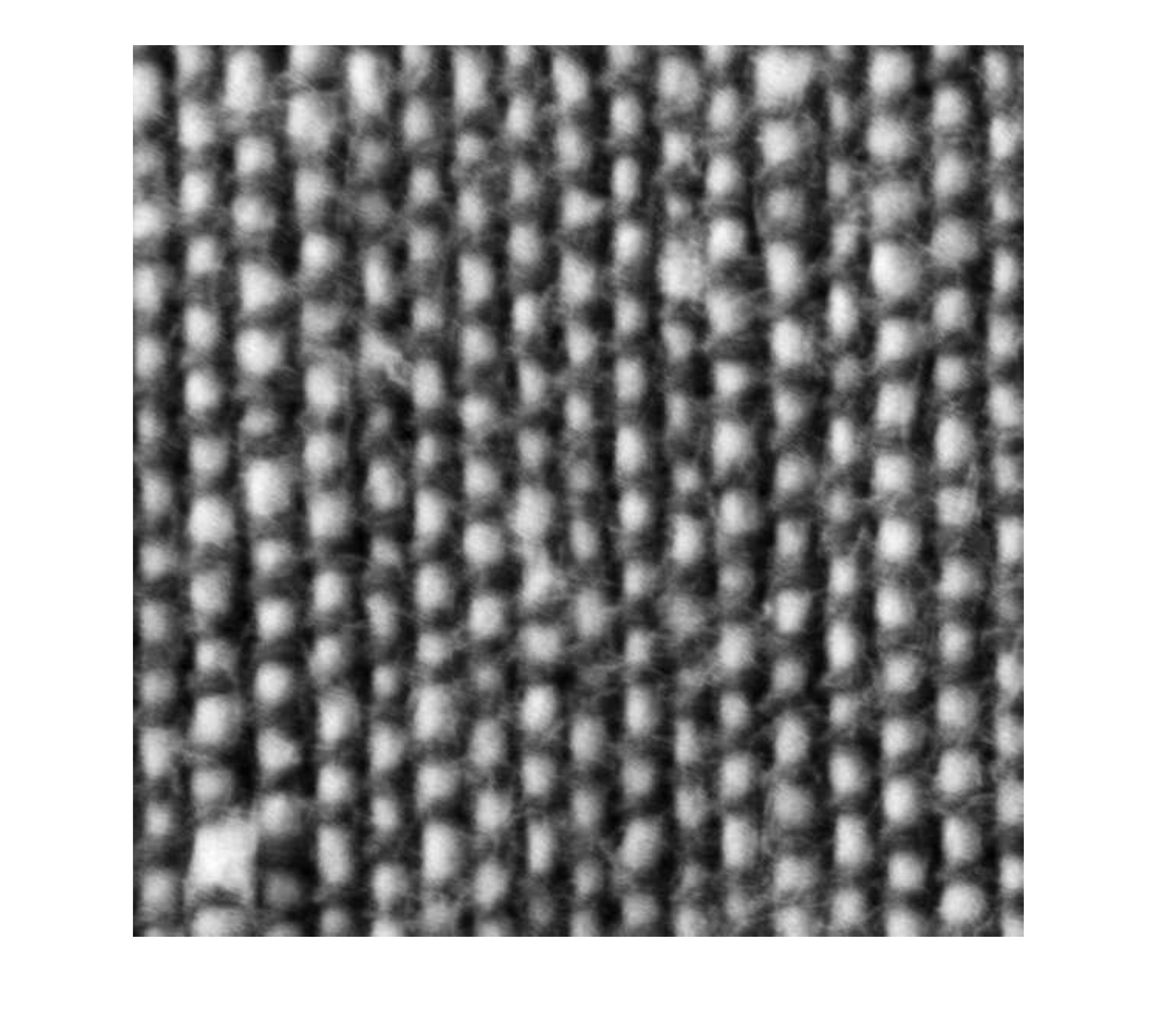}}
				{(d)}\medskip
			\end{minipage}\hspace{0.01mm}
			\begin{minipage}[b]{0.31\linewidth}
				\centering
				{\includegraphics[width=4cm]{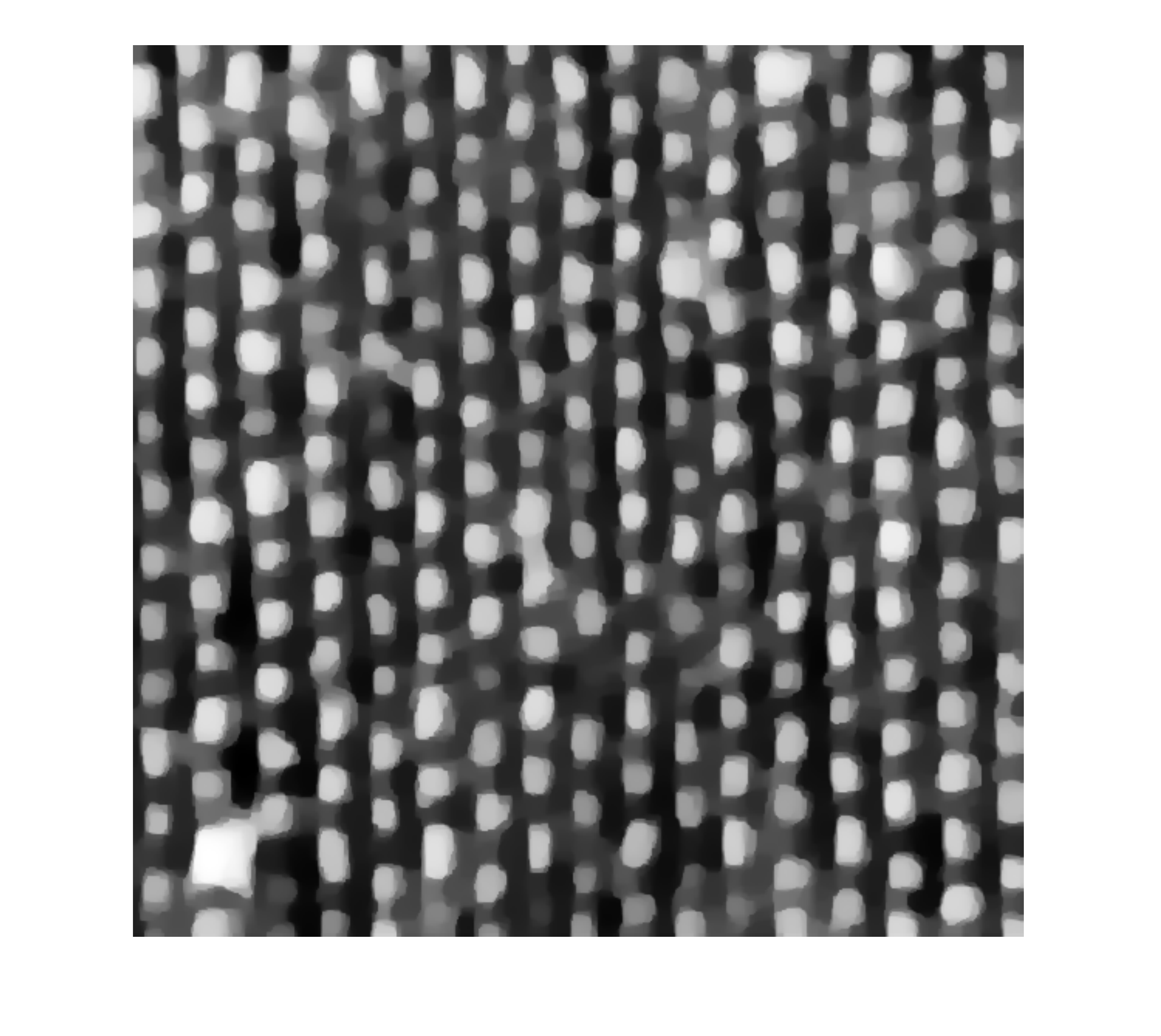}}
				{(e)}\medskip
			\end{minipage}
			\hspace{0.1mm}
			\begin{minipage}[b]{0.31\linewidth}
				\centering
				{\includegraphics[width=4cm]{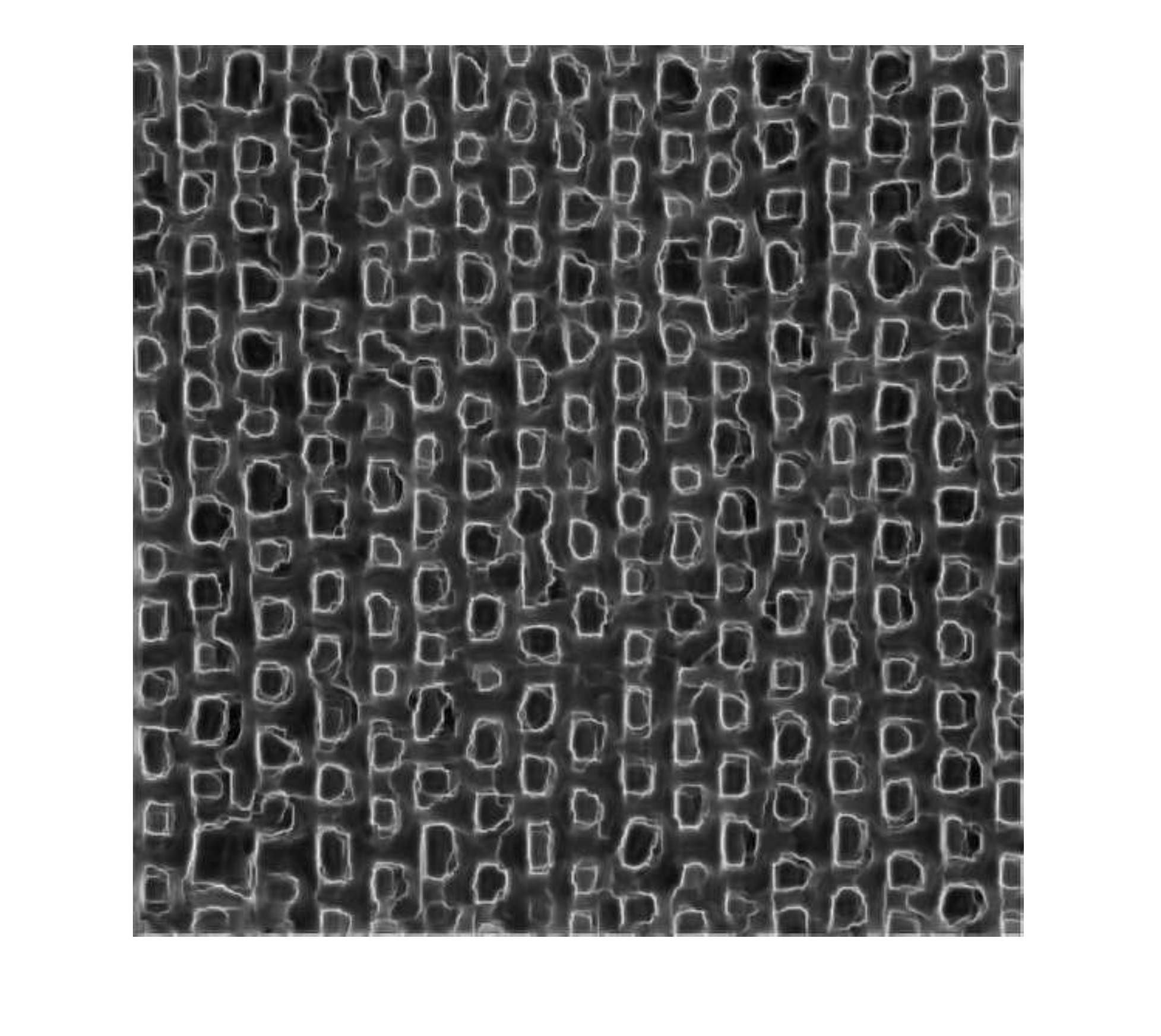}}
				{(f)}\medskip
			\end{minipage}
			\caption{Examples of the 2D phase congruency (PC) extracted from structural part of Kylbreg images. (a,d) Examples of images belonging to different classes (linseeds and cushion). (b,e) Their corresponding structural layer extracted using the RTV method. (c,f) The 2D PC, which convey edge-like information. }\label{fig:2Dpc}
		\end{figure}

\begin{figure}[t]
		\begin{minipage}[b]{0.21\linewidth}
			\centering
			{\includegraphics[width=4cm]{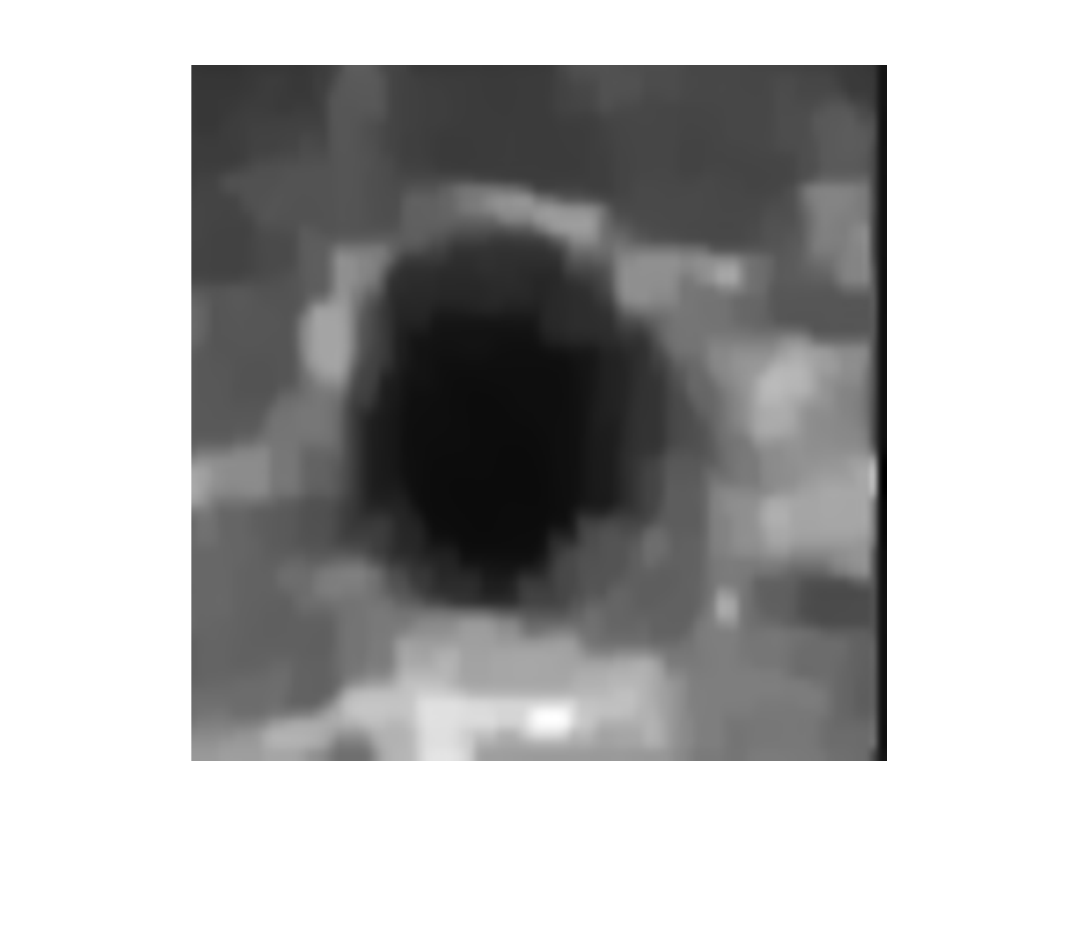}}
			{(a)}\medskip
		\end{minipage}\hspace{0.01mm}
		\begin{minipage}[b]{0.21\linewidth}
			\centering
			{\includegraphics[width=4cm]{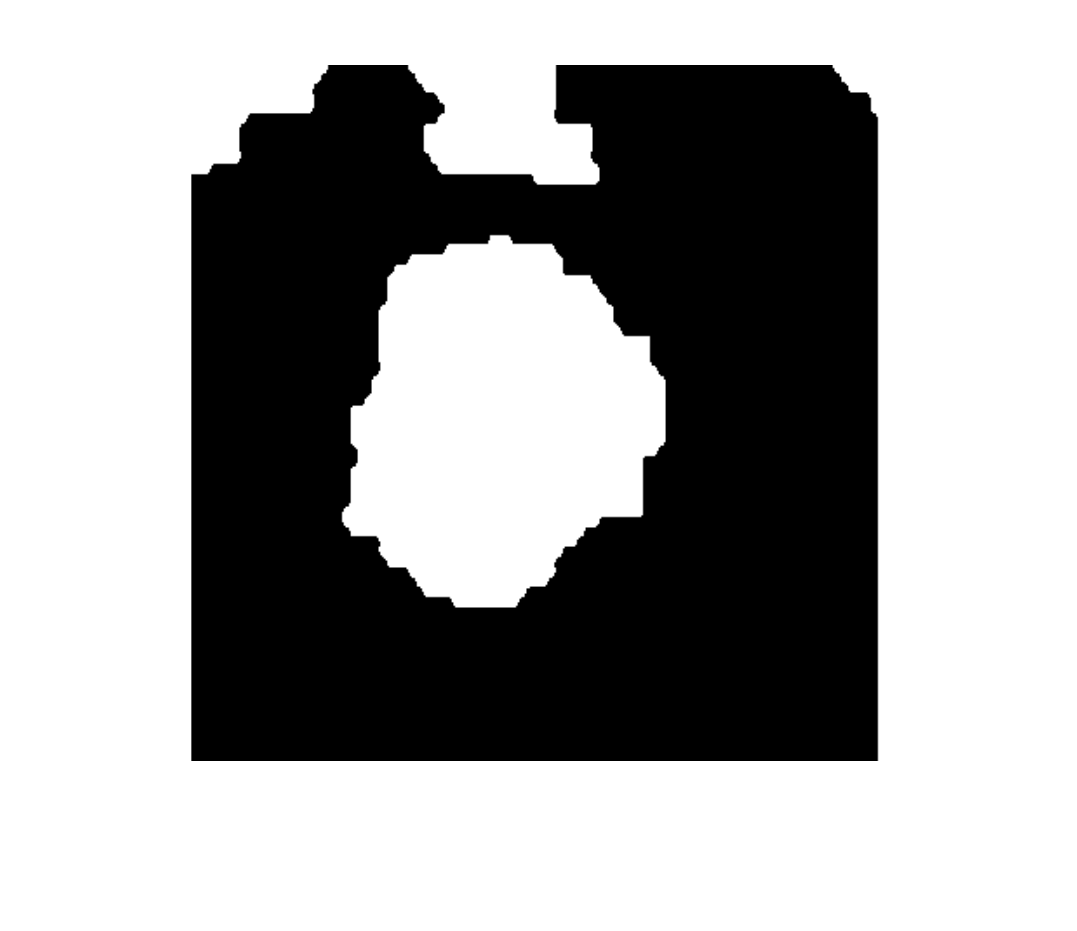}}
			{(b)}\medskip
		\end{minipage}
		\begin{minipage}[b]{0.21\linewidth}
			\centering
			{\includegraphics[width=4cm]{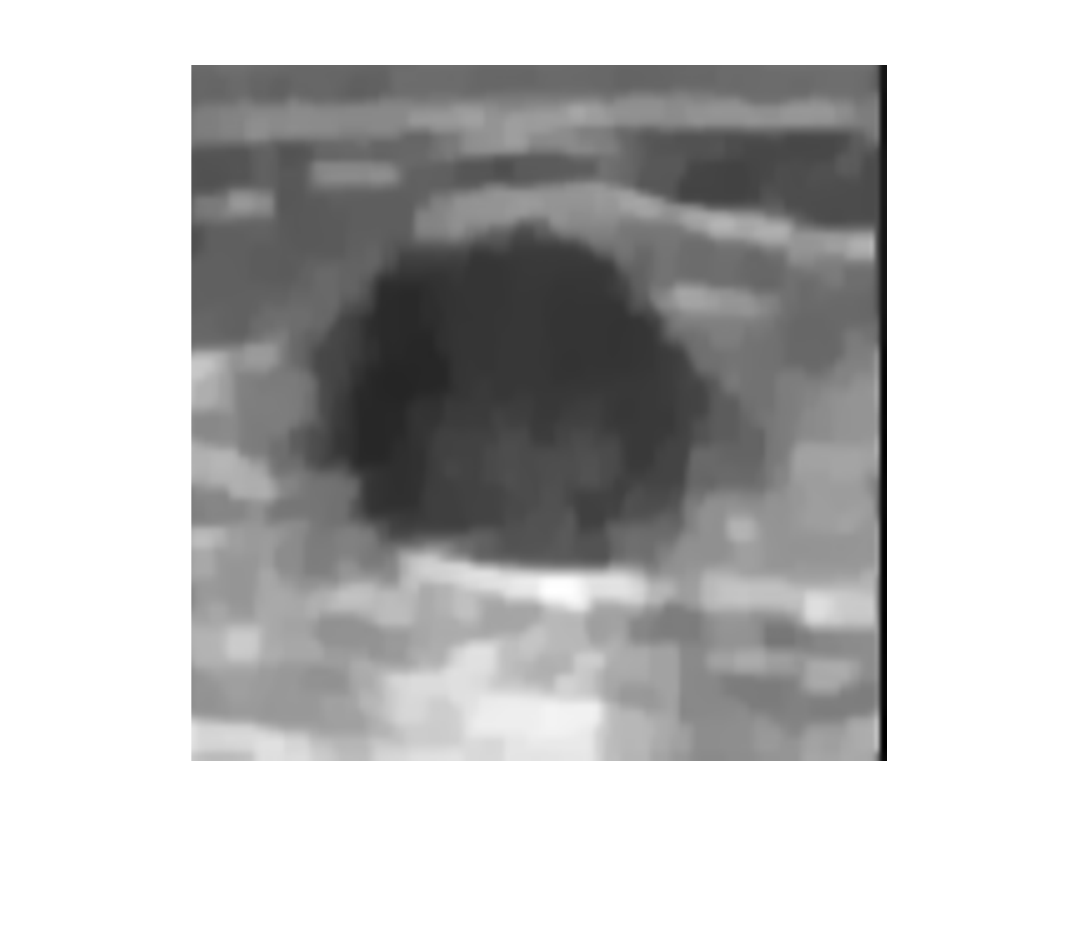}}
			{(c)}\medskip
		\end{minipage}
		\hspace{0.1mm}
		\begin{minipage}[b]{0.21\linewidth}
			\centering
			{\includegraphics[width=4cm]{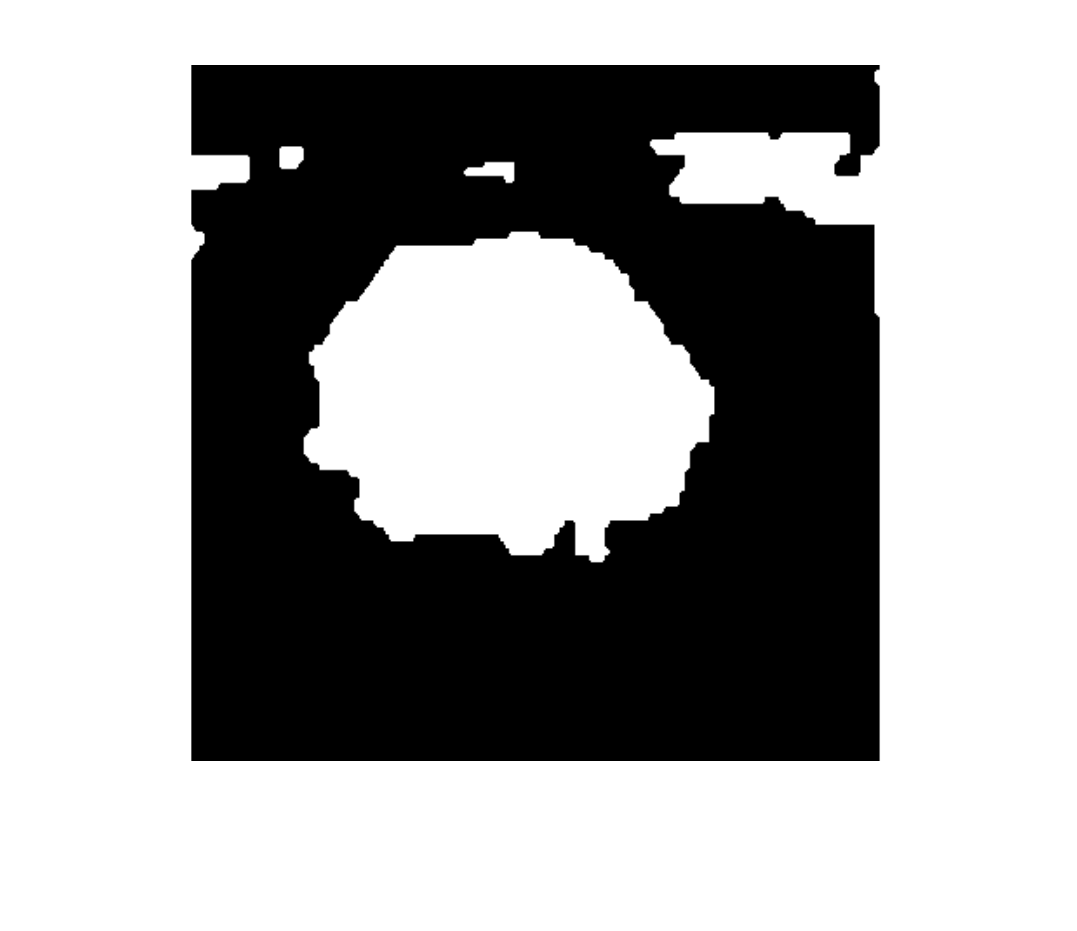}}
			{(d)}\medskip
		\end{minipage}
		\caption{Examples of binarization of BUSIS structural images for the purpose of assessing tumor size. (a,c) Structure images of B and M cases, respectively. (b,d) Their corresponding binary mask obtained by STH algorithm, as described in Alg.\ref{Alg:1}.}\label{fig:BUSISbin}
	\end{figure}

		Structural information or edge-like features vary from image to another, and it isn't easy to quantify it by means of a single parameter, similarly to the case of NST. We therefore extract different structural features for different classes of images. In the case od BUSIS images, the structural features related to the tumor size, area and structure, whereas, in the case of the Kylbreg images, we quantify the structure by modeling the local phase using the quantifier of Phase congruency (PC).\\
		\indent The PC, or other measures of phase, provide a good measure of edge-like features in images such as lines, edges and contours \cite{kovesi1999image,behar1992image}. We use the modified PC measure \cite{kovesi1999image} to quantify the local phase information: 
		\begin{equation}
		PC(x)=\frac{\sum_{n}W(x)\left\lfloor A_{n}(x)(\zeta(x))-\gamma\right\rfloor }{\sum_{n}A_{n}(x)+\epsilon},
		\end{equation}
		where $\zeta(x)=\cos(\varphi_{n}(x)-\bar{\varphi}(x))-|\sin(\varphi_{n}(x)-\bar{\varphi}(x)|$, $W(x)$ is a frequency weight factor, $A_{n}(x)$ and $\varphi_{n}(x)$ are the amplitude and phase in the wavelets' n-level, respectively. $\gamma$ is a noise threshold that renders the PC measure to become robust with respect to noise. The operator $\left\lfloor \cdot\right\rfloor$ denotes that the quantity equals to itself when it is positive and it is zero otherwise. We calculate the 2D PC of the structural layer of kylbreg images (See examples of the 2D PC in Fig.\ref{fig:2Dpc}) and use the mean of PC as an additional feature \cite{PC3}.\\
		\indent In the case of BUSIS images, we take advantage of features related to the tumor structure. As one can notice in the structural part of the BUSIS image, in Fig.\ref{fig:layers}(f), the area of the tumor is dominated by pixels of low gray-levels. We therefore easily define this area by thesholding the structure image, by keeping only the low-gray-level pixels. By thresholding and labeling we will keep also other objects around the tumor, which are falsely-detected objects. Indeed, the tumor region is located at the center of the ROI. Therefore, we can filter the other falsely-detected objects by calculating their distance from the center. In light of the aforementioned facts, we propose the structure thresholding (STH) algorithm, described in Algo.\ref{Alg:1}, that estimate the area of the tumor. It appears that geometrical measures can significantly contribute in B and M detection. Here, we proposed a straightforward quantifier. Alternately, measures that related to the contour curvature and its 1D signatures are also significant for characterizing the tumor shape \cite{rangayyan2007fractal,nguyen2006shape}. Most of tumors that appear in the BUSIS images exhibit almost similar shapes, but with different geometrical features (size and/or area). This fact leads us to estimate their area and use it as a straightforward quantifier. However, in more complex cases, such encountered in mammography, tumors may appear with different shapes. The latter demands extraction of shape-related features such as curvatures of the 2D tumor segment or its 1D contour. In fact, the robustness of these calculation affected by the resolution of the images. In our case, the BUSIS images are not of sufficient resolution that allows curvature extraction.\\
		\indent  Examples of performing the STH process on the BUSIS dataset are presented in Fig.\ref{fig:BUSISbin}. As we see, it is hard by eyeballing to distinguish between the shapes of B and M tumors. STH should be efficient in this task, in that it provides a quantitative measure of the tumor size. In our classification task, we employ the area calculated from STH as a structural feature. 
		\begin{algorithm}[t]
			\SetKwInOut{Input}{Input}
			\SetKwInOut{Output}{Output}		
			\Input{BUS Structure Image $S$}
			\Output{Area}	
			\indent Apply histogram equalization on $S$. The resultant image is $S_{eq}$.\\
	  		\indent Quantize $S_{eq}$ to obtain $\tilde{S}$ with 5 levels: 
			  $l=\left\{ 0,1,..,4\right\}$. \\ 
	 		\indent Construct binary mask from $\tilde{S}$ as follows: 			$B(i,j)=1$ if $\tilde{S}\left(i,j\right)<3$ and $0$ otherwise.
	
			Label the objects in $B$ and find their centers \\
			Find the object with minimal Euclidean distance from the ROI's center and calculate its area, $A$.\\
		    Return $A$	
			\caption{Structure Thresholding (STH) Algorithm}\label{Alg:1}
		\end{algorithm}
	\subsection{Fusion of Experts}\label{sec:fusion}
	Features extracted from texture and structure are treated as multi-view sets of features. Two SVMs are accordingly trained on them independently. We then merge between the  SVMs outputs by training a shallow NN, which optimizes the overall performance. In practice, the NN should be trained on examples that are not included in the SVMs training sets, in order to seek a generalized optimal decision.\\
	\indent We begin by dividing our two datasets (BUSIS and Kylbreg) into three subsets of data: the first one to be used in training the SVMs, the second for training the NN, and the third one for testing the performance of the  system as a whole. We train two SVMs on corresponding structural, $\Phi_{S}$, and textural, $\Phi_{T}$, features extracted from the training set. Then, for each example in the NN training set, the distances from the hyperplane of structural and textural SVMs, $d_{T}$ and $d_{S}$, are calculated. Lastly, the NN is trained on the vector $d_{T}\oplus d_{S}$ obtained by concatenating the S and T distances corresponding the two-view features. In the case of binary classification, each $d_{S}$ and $d_{T}$ are one-dimensional vectors, whereas, in multi class classification, $k(k-1)/2$ binary classifiers are trained for the $k$ multi-class task \cite{bishop2006pattern}. The dimension of the resultant vector is $k(k-1)$.\\
	
	\textit{The NN Architecture:}
		As our database does not include large amount of examples, the NN should be shallow, to avoid over-fitting. We therefore suggest the implementation of NN with only three fully connected layers using ReLu activation function. The first layer is of size $k(k-1)\times8$, the second is of $8\times4$ and the last layer (output layer) consists of $k$ neurons, where $k$ is the number of classes ($k=2$ in the case of BUSIS dataset and $k=6$ in the case of Kylbreg). 		
	\subsection{Experimental Results and Discussion}\label{sec:exp}
	 \paragraph{Dataset:}\mbox{}\\
	  \indent Our experiments are performed on  the Kylbreg textures database \cite{Kylbreg2011c}. This dataset consists of $240$ fully-textured images belonging to $6$ different substances: sand, seeds, canvas, cushion, seat and stone. We also demonstrate the applicability of our proposed method on  the public breast ultrasound (BUSIS) dataset \cite{xian2018benchmark}, which includes images of benign (B) and malignant (M) cases.
	\begin{table}[b]
		\caption{Classification results: accuracy obtained with Kylbreg (top) and BUSIS (down) datasets. Classification results obtained by our two-view system, denoted by $d_{T}\oplus d_{S}$, are compared with three other modes: $T$ and $S$ stand for using SVM trained on only one view, i.e either textural or structural features, respectively. $T\oplus S$ stands for SVM trained on a two-view set, created by concatenating the features $\Phi_{S}$ and $\Phi_{S}$. For both datasets, our two-view approach boosts the accuracy in classification of the test data.}\label{table:1}
		\begin{minipage}{.9\linewidth}
			\centering
			\begin{tabular}{|c|c|c|c|c|}
				\hline 
				& $\mathbf{T}$ & $\mathbf{S}$ & $\mathbf{T\oplus S}$ & \textbf{$\mathbf{d_{T}\oplus d_{S}}$}\tabularnewline
				\hline 
				\textbf{Test} & $76.2\%$ & $81.2\%$ & $94\%$ & $95.7\%$\tabularnewline
				\hline 
			\end{tabular}
		\end{minipage}\par
		\vspace{0.2cm}
		\begin{minipage}{.9\linewidth}
			\centering
			\begin{tabular}{|c|c|c|c|c|}
				\hline 
				& $\mathbf{T}$ & $\mathbf{S}$ & $\mathbf{T\oplus S}$ & \textbf{$\mathbf{d_{T}\oplus d_{S}}$}\tabularnewline
				\hline 
				\textbf{Test} & $82.5\%$ & $88.0\%$ & $88.2\%$ & $91.0\%$\tabularnewline
				\hline 
			\end{tabular}
		\end{minipage} 
	\end{table}
	\begin{figure}[t]
	\begin{minipage}[b]{0.41\linewidth}
		\centering
		{\includegraphics[width=6cm]{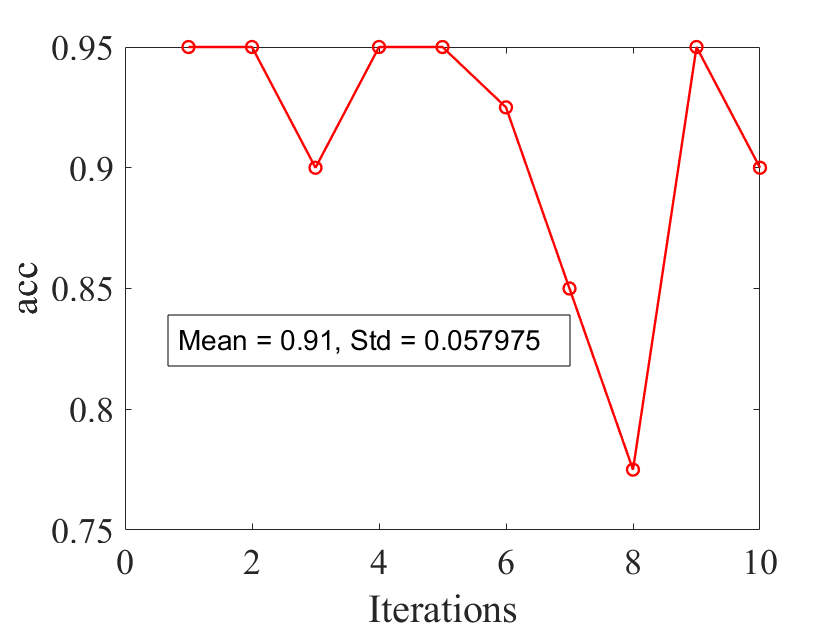}}
		{(a)}\medskip
	\end{minipage}\hspace{0.8cm}
	\begin{minipage}[b]{0.41\linewidth}
		\centering
		{\includegraphics[width=6cm]{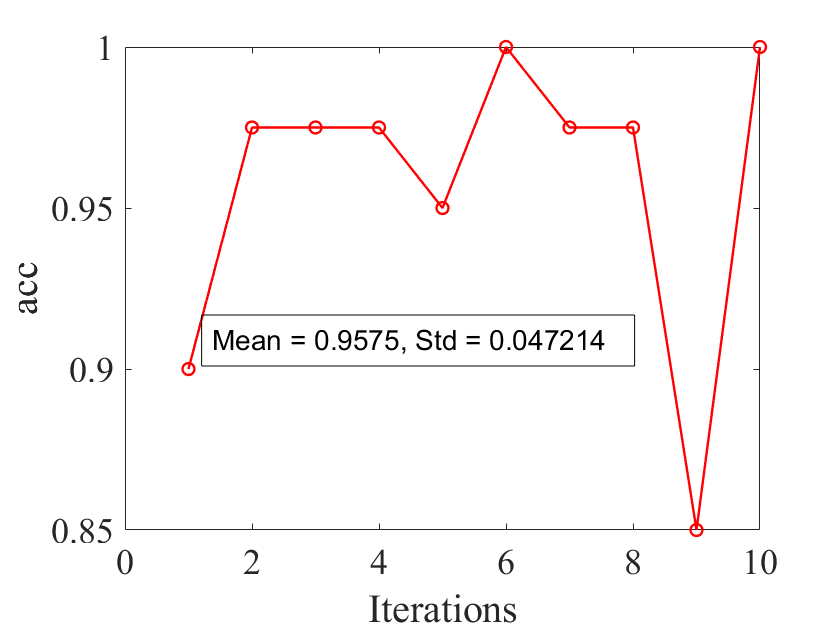}}
		{(b)}\medskip
	\end{minipage}
	\caption[Classification performance with Repetition]{The accuracy of the classification as a function of iterations, obtained with BUSIS (a) and Klybreg (b) datasets, after performing 10 repetitions.}\label{fig:accrep}
	\end{figure} 
	  \paragraph{Kylbreg Textures Classification:}\mbox{}\\
	  \indent We divide the textured image into $32\times32$ non-overlapping patches and estimate their Hurst parameter (H). The mean and variance of the estimated H are utilized as textural features, $\Phi_{T}$. From the structural part, we extract the 2D PC and use its mean as the complementary structural feature, $\Phi_{S}$. Datasets are divided randomly into 3 groups: half of the examples are utilized for training the SVMs, $40$ examples are used for test and the remaining data are saved for training the shallow NN. We then train two multi-class SVMs with radial basis function kernel (RBF), applied on  textural and structural features independently, using the LIBSVM implementation \cite{chang2011libsvm}. Then, distances from structural and textural hyperplanes are concatenated and used in training the NN (as outlined in section \ref{sec:fusion}) to optimize the overall decision making process. 
	  \paragraph{BUSIS images Classification:}\mbox{}\\ 
	 \indent The same training procedure is also performed in the case of BUSIS images. But, here the structural features, $\Phi_{S}$ are related to the tumor's 
	  structural properties, as clarified in section \ref{sec:feat}. The $\Phi_{S}$ are extracted using the STH algorithm, described in Algo.\ref{Alg:1}.\\	  
	  \indent Results of our classification method on the Kylbreg (left) and BUSIS (right) datasets are presented in Table.\ref{table:1}. Note that these results are the mean accuracy calculated after repeating  the training and testing process 10 iterations, where at each iteration the test and train sets were chosen randomly. We performed these repetitions in order to get a robust estimate of the classification performance. Results of the accuracy as a function of repetition number are highlighted in Fig.\ref{fig:accrep}.
	   As one can conclude from the comparison made in Table.\ref{table:1}, the accuracy of our two-view approach, denoted by $d_{T}\oplus d_{S}$, achieves the best accuracy compared with using only one set of the features (columns $S$ and $T$), or by concatenating between them. Table \ref{table:2} presents head-to-head comparison between our two-view method classification results and the performance recently reported in \cite{lee2018intensity}. In latter, the authors propose the use of intensity inhomogeneity correction and stacked denoising autoencoder (SADE) is proposed for BUSIS classification. Due to the small amount of data, the SADE yields the best performance compared with other deep learning (DNN) methods. In Table \ref{table:2}, we compare our method with the SADE and the four other methods mentioned in \cite{lee2018intensity}. As indicated by the results depicted in the first columns, our proposed two-view-based method yields the best classification performance.\\
	  
\begin{figure}[t] 
	  	\begin{minipage}[b]{0.5\linewidth}
	  		\centering
	  		{\includegraphics[width=14cm]{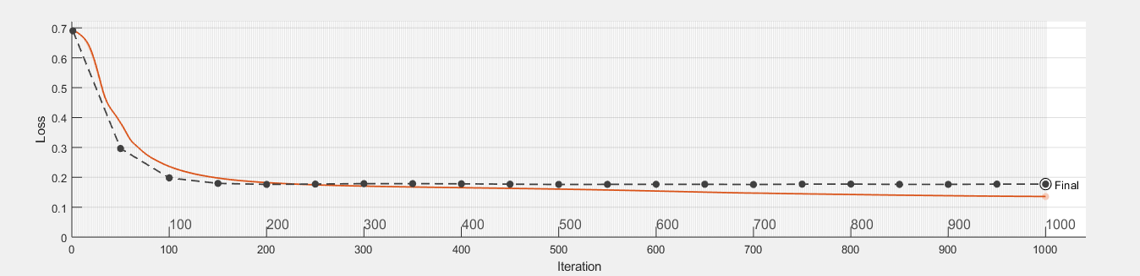}}
	  		{(a)}\medskip
	  	\end{minipage}\hspace{0.01mm}\\
	  	\begin{minipage}[b]{0.5\linewidth}
	  		\centering
	  		{\includegraphics[width=14cm]{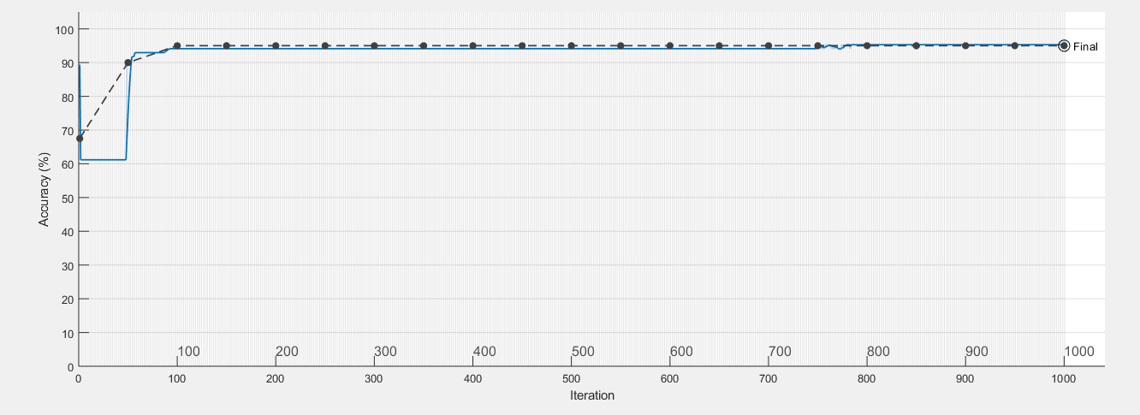}}
	  		{(b)}\medskip
	  	\end{minipage}\\
	  	\begin{minipage}[b]{0.5\linewidth}
	  		\centering
	  		{\includegraphics[width=14cm]{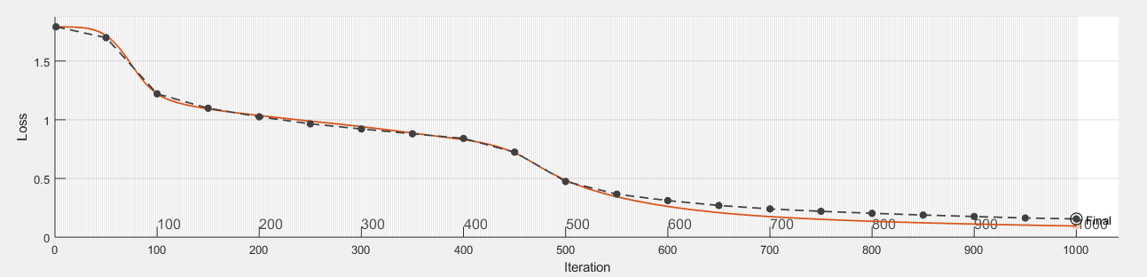}}
	  		{(c)}\medskip
	  	\end{minipage}\hspace{0.01mm}\\
	  	\begin{minipage}[b]{0.5\linewidth}
	  		\centering
	  		{\includegraphics[width=14cm]{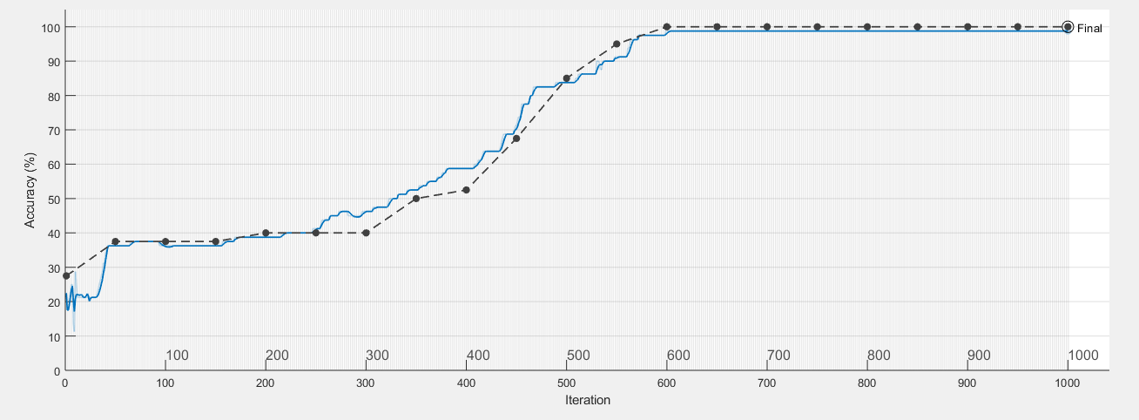}}
	  		{(d)}\medskip
	  	\end{minipage}
	  	\caption{Loss and accuracy of the NN as a function of the number of epochs (iterations) in the case of BUSIS images (a,b) and of Kylbreg images (c,d) classification. Test Loss and accuracy are highlighted in dashed-black, training loss is plotted with solid-orange and training accuracy appears in solid-blue.  }\label{fig:lossacc}
	  \end{figure}

	  \indent In both classification problems, $1000$ epochs were needed to decrease the training loss almost to zero (See curves of training loss and accuracy as a function of epochs number in Fig.\ref{fig:lossacc}). Since we use a NN with very few hidden layers, the training loss decreases with epochs and the training process is relatively fast. Further, the classification loss converges to its optimal value relatively rapidly, for both test and train subsets.\\
	
\section{Conclusions}
Analysis of natural textures that represent various materials, either encountered in medical images (such as Breast ultrasound or even mammography), or represented by images acquired from in other natural substances, brings
to the conclusion that in a large number of uniformly distributed
NST there exists some structural information that is reflected by the
presence of coherent local spatial phase. Consequently, even
images of uniformly distributed natural materials must be first separated into the two layers of the structured/deterministic (i.e. phase-based) and random/fractal (i.e fBm-based)
components of the image,  before being represented for classification in a combined structured-textured space.
 Here we proposed the two-view method, exploiting both fractal (textural-related) and structural features. Combining both becomes of paramount importance when more structural (edge-and-contour-type) information is present in the natural images. The classification method was constructed in this paper in a straightforward manner, by fusing the "decisions" of two independent SVMs, trained in the structural and textural spaces, respectively. Another, more challenging approach to merge between the two aforementioned views, is by embedding the non-canonical features in a suitable manifold that lends itself to a geometric representation that comes along with a natural metric \cite{sochen1998representation}.
\begin{table}[t]
	\caption{Comparison between the proposed two-view method, the SADE and four other DNN methods mentioned in \cite{lee2018intensity}. The performance of SADE and the DNN are presented as: $X/Y$, where $X$ is the obtained result after correction for inhomogeneity and $Y$ denotes the result before correction.}\label{table:2}
	\centering
	\scalebox{0.92}{
	\begin{tabular}{|c|c|c|c|c|c|c|}
		\hline 
		& \textbf{Ours} & \textbf{SDAE} & \textbf{AlexNet} & \textbf{Inception v3} & \textbf{ResNet} & \textbf{DenseNet}\tabularnewline
		\hline 
		\hline 
		\textbf{Precision } & \textbf{$0.88$} & $0.71/0.78$ & $0.56/0.73$ & \textbf{$0.68/0.79$} & $0.69/0.61$ & $0.71/0.71$\tabularnewline
		\hline 
		\textbf{Recall } & $0.88$ & $0.85/0.9$ & $0.5/0.8$ & $0.65/0.75$ & $0.8/0.8$ & $0.75/0.85$\tabularnewline
		\hline 
		\textbf{Specificity } & $0.91$ & $0.65/0.75$ & $0.6/0.7$ & $0.7/0.8$ & $0.65/0.5$ & $0.7/0.65$\tabularnewline
		\hline 
		\textbf{F-measure} & $0.87$ & \multicolumn{1}{c|}{$0.77/0.83$} & $0.53/0.76$ & $0.67/0.77$ & $0.74/0.70$ & $0.74/0.70$\tabularnewline
		\hline 
		\textbf{Accuracy} & $0.91$ & 0$.75/0.83$  & $0.55/0.75$ & $0.68/0.78$ & $0.73/0.65$ & $0.73/0.75$\tabularnewline
		\hline 
	\end{tabular}}
\end{table}

\section{Mathematical Proofs}\label{mathproofs}
\subsection{Proof of Proposition 1}
\begin{proof}
	proposition.\ref{prop1} can be directly proven by showing the invariance of the two-first moments along scales, or levels $j$, based on the Gaussianity of the multiresolution pyramid denoted by $c_{j,k}$. Assuming  $\E(B_{H}(t))=0$  $\forall t\in\mathbb{R}$, we obtain the equality in the variance:
	$ $
	\[
	\text{Var(\ensuremath{c_{j-1,k}})}=\E[c_{j-1,k}^{2}]=\E\left[(2^{j-1}\intop_{-\infty}^{\infty}B_{H}(t)\Psi(2^{j-1}t-k)\,dt)^{2}\right]
	\]
	\[
	=2^{2(j-1)}\E\left[\intop_{-\infty}^{\infty}\intop_{-\infty}^{\infty}B_{H}(t)B_{H}(s)\Psi(2^{j-1}s-k)\Psi(2^{j-1}t-k)\,dsdt\right]
	\]
	\[
	\underset{\textrm{Linearity\,of\,}\E}{=}\frac{2^{2j}}{4}\intop_{-\infty}^{\infty}\intop_{-\infty}^{\infty}\E\ensuremath{[B_{H}(t)B_{H}(s)]}\ensuremath{\Psi(2^{j}\frac{s}{2}-k)\Psi(2^{j}\frac{t}{2}-k)\,dsdt}
	\]
	\[
	\underset{\widetilde{t}=t/2,\widetilde{s}=s/2}{=}\frac{2^{2j}}{4}\intop_{-\infty}^{\infty}\intop_{-\infty}^{\infty}\E[B_{H}(\tilde{t}/2)B_{H}(\tilde{s}/2)] \ensuremath{\Psi(2^{j}\widetilde{t}-k)\Psi(2^{j}\widetilde{s}-k)\cdot4\,d\widetilde{s}d\widetilde{t}}
	\]
	\[
	\underset{(*)}{=}2^{2j}\intop_{-\infty}^{\infty}\intop_{-\infty}^{\infty}\frac{2^{2H}}{2}\left(|\widetilde{t}|^{2H}+|\widetilde{s}|^{2H}+|\widetilde{t}-\widetilde{s}|^{2H}\right)\ensuremath{\Psi(2^{j}\widetilde{t}-k)\Psi(2^{j}\widetilde{s}-k)\,d\widetilde{s}d\widetilde{t}}
	\]
	\[
	=2^{2j}E\left[\intop_{-\infty}^{\infty}\intop_{-\infty}^{\infty}2^{H}B_{H}(t)\Psi(2^{j}s-k)2^{2H}B_{H}(s)\Psi(2^{j}t-k)\,dsdt\right]=\text{Var(\ensuremath{2^{H}c_{j,k}})}
	\]
	$(*)$ This equality is valid because of the self-similarity of $B_{H}(t)$ i.e equality of the covariances: 
	\[
	\E\left[B_{H}(\alpha t)B_{H}(\alpha s)\right]=\frac{1}{2}\left(||\alpha t||^{2H}+||\alpha s||^{2H}-||\alpha(t-s)||^{2H}\right)
	\]	\[
	=\frac{|\alpha|^{2H}}{2}\left(||t||^{2H}+||s||^{2H}-||t-s||^{2H}\right)= \E\left[B_{H}(t)B_{H}(s)\right]
	\]
\end{proof}
\indent In the proof we consider the 1D fractional Brownian motion (fBm),i.e. $t\in\mathbb{R}$. However, it can be directly extended to the 2D fBm,  by replacing the absolute value in $(*)$ (in the auotocorrlation of fBm) with the Euclidean norm operator $||\cdot||$. 
\subsection{Proof of Proposition 2}
\begin{proof}
	The ML estimators $\hat{\sigma}_{1}$and $\hat{\sigma}_{2}$ are consistent,
	i.e. they converges to the real parameters when the n-samples are
	large enough. Due to the self-similarity of $c_{j,k}$ and $2^{-H}c_{j-1,k}$,
	$\hat{\sigma}_{1}$ and $\hat{\sigma}_{2}$ converge to the same parameter, i.e.\\ $\lim_{n\rightarrow\infty}\hat{\sigma}_{2}-\hat{\sigma}_{1}=0$. Therefore. there exists $N$ s.t $\forall n>N:\:|\hat{\sigma}_{1}-\hat{\sigma}_{2}|\leq\epsilon$.
	The KL diveregence between the two PDFs 
	of $c_{j,k}$ and $2^{-H}c_{j-1,k}$, $P_{1}$ and $P_{2}$, is given
	by: 
	\[
	D_{KL}(p_{1}||p_{2})=\log\left(\frac{\hat{\sigma}_{2}}{\hat{\sigma}_{1}}\right)+\frac{\hat{\sigma}_{1}^{2}}{2\hat{\sigma}_{2}^{2}}-\frac{1}{2}
	\]
	
	$\hat{\sigma}_{2}-\epsilon\leq\hat{\sigma}_{1}\leq\hat{\sigma}_{2}+\epsilon$
	. Therefore: 
	\[
	D_{KL}(p_{1}||p_{2})<\log(\frac{\hat{\sigma}_{2}}{\hat{\sigma}_{2}-\epsilon})+\frac{(\hat{\sigma}_{2}+\epsilon)^{2}}{2\hat{\sigma}_{2}^{2}}-\frac{1}{2}
	\]
	\[
	=-\log(1-\frac{\epsilon}{\hat{\sigma}_{2}})+\frac{1}{2}+\frac{\epsilon}{\hat{\sigma}_{2}}+\frac{1}{2}(\frac{\epsilon}{\hat{\sigma}_{2}})^{2}-\frac{1}{2}\underset{(**)}{\cong}\frac{\epsilon}{\hat{\sigma}_{2}}+\text{\O\ensuremath{(\frac{\epsilon}{\hat{\sigma}_{2}})^{2}+(\frac{\epsilon}{\hat{\sigma}_{2}})^{2}}}=O(\frac{\epsilon}{\hat{\sigma}_{2}}).
	\]
	\indent $(**)$ This equality results from first order Taylor expansion of $\log(1-x)$, as $\epsilon\ll\hat{\sigma}_{2}$:\\
	$\log(1-\frac{\epsilon}{\hat{\sigma}_{2}})\simeq\frac{\epsilon}{\hat{\sigma}_{2}}+\text{\O(\ensuremath{\frac{\epsilon}{\hat{\sigma}_{2}})^{2}}}$, where ${\O}(\frac{\epsilon}{\hat{\sigma}_{2}})^{2}$  is the residual or the approximation error.\\	
\end{proof}
\newpage
\footnotesize
\bibliographystyle{splncs03}
\bibliography{egbib}

\end{document}